%% file: main.tex
\documentclass{article}

\usepackage{microtype}
\usepackage{graphicx}
\usepackage{subfigure}
\usepackage{booktabs} 

\usepackage{hyperref}

\usepackage[accepted]{icml2021}

\icmltitlerunning{Simultaneous Similarity-based Self-Distillation for Deep Metric Learning}

\usepackage{amsmath,amssymb} 

\usepackage{microtype}
\usepackage{graphicx}
\usepackage{comment}
\usepackage{color}
\usepackage[utf8]{inputenc} 
\usepackage{url}            
\usepackage{booktabs}       
\usepackage{amsfonts}       
\usepackage{nicefrac}       
\usepackage{microtype}      
\usepackage{color}

\usepackage{listings}
\usepackage{xcolor,colortbl}
\definecolor{codegreen}{rgb}{0,0.6,0}
\definecolor{codegray}{rgb}{0.5,0.5,0.5}
\definecolor{codepurple}{rgb}{0.58,0,0.82}
\definecolor{backcolour}{rgb}{0.95,0.95,0.95}

\lstdefinestyle{mystyle}{
    backgroundcolor=\color{backcolour},   
    commentstyle=\color{codegreen},
    keywordstyle=\color{magenta},
    numberstyle=\tiny\color{codegray},
    stringstyle=\color{codepurple},
    basicstyle=\ttfamily\footnotesize,
    breakatwhitespace=false,         
    breaklines=true,                 
    captionpos=b,                    
    keepspaces=true,                 
    numbers=left,                    
    numbersep=5pt,                  
    showspaces=false,                
    showstringspaces=false,
    showtabs=false,                  
    tabsize=2
}

\definecolor{vvlightgray}{rgb}{0.9,0.9,0.9}
\definecolor{vlightgray}{rgb}{0.8,0.8,0.8}

\usepackage{wrapfig}
\graphicspath{ {./images/} }
\usepackage{multirow}
\usepackage{sidecap}

\newcommand{\blue}[1]{\textcolor{blue}{#1}}

\DeclareMathOperator*{\argmin}{arg\,min}
\DeclareMathOperator*{\minsort}{min\,sort}

\begin{document}

\twocolumn[
\icmltitle{Simultaneous Similarity-based Self-Distillation for Deep Metric Learning}



\icmlsetsymbol{equal}{*}

\begin{icmlauthorlist}
\icmlauthor{Karsten Roth}{ut,hd}
\icmlauthor{Timo Milbich}{hd}
\icmlauthor{Björn Ommer}{hd}
\icmlauthor{Joseph Paul Cohen}{equal,mil}
\icmlauthor{Marzyeh Ghassemi}{equal,mit,ut}
\end{icmlauthorlist}

\icmlaffiliation{mit}{MIT}
\icmlaffiliation{ut}{University of Toronto, Vector Institute}
\icmlaffiliation{hd}{Heidelberg University, IWR}
\icmlaffiliation{mil}{Mila, Université de Montréal}

\icmlcorrespondingauthor{Karsten Roth}{karsten.rh1@gmail.com}

\icmlkeywords{Machine Learning, ICML}

\vskip 0.3in
]



\printAffiliationsAndNotice{\icmlEqualContribution} 

\begin{abstract}
Deep Metric Learning (DML) provides a crucial tool for visual similarity and zero-shot applications by learning generalizing embedding spaces, although recent work in DML has shown strong performance saturation across training objectives.
However, generalization capacity is known to scale with the embedding space dimensionality. Unfortunately, high dimensional embeddings also create higher retrieval cost for downstream applications.
To remedy this, we propose \emph{Simultaneous Similarity-based Self-distillation (S2SD)}.  \textit{S2SD} extends DML with knowledge distillation from auxiliary, high-dimensional embedding and feature spaces to leverage complementary context during training while retaining test-time cost and with negligible changes to the training time. 
Experiments and ablations across different objectives and standard benchmarks show \textit{S2SD} offers notable improvements of up to 7\% in Recall@1, while also setting a new state-of-the-art.
Code available at \url{https://github.com/MLforHealth/S2SD}.
\end{abstract}

\section{Introduction}
Deep Metric Learning (\textit{DML}) aims to learn embedding space models in which a predefined distance metric reflects not only the semantic similarities between training samples, but also transfers to unseen classes. The generalization capabilities of these models are important for applications in image retrieval \citep{margin}, face recognition \citep{semihard}, clustering \citep{grouping} and representation learning \citep{moco}.
Still, transfer learning into unknown test distributions remains an open problem, with \citet{roth2020revisiting} and \citet{musgrave2020metric} revealing strong performance saturation across DML training objectives. However, 
\citet{roth2020revisiting} also show that embedding space dimensionality can be a driver for generalization across objectives due to higher representation capacity. 
Indeed, this insight can be linked to recent work targeting other objective-independent improvements to DML via artificial samples \citep{hardness-aware}, higher feature distribution moments \citep{horde} or orthogonal features \citep{milbich2020diva}, which have shown promising relative improvements over selected DML objectives.
Unfortunately, these methods come at a cost; be it longer training times or limited applicability.
Similarly, drawbacks can be found when naively increasing the operating (\textit{base}) dimensionality, incurring increased cost for data retrieval at test time, which is especially problematic on larger datasets. This limits realistically usable embedding dimensionalities and leads to benchmarks being evaluated against fixed, predefined dimensionalities.\\
In this work, we propose \textit{Simultaneous Similarity-based Self-Distillation} (\textit{S2SD}) to show that complex higher-dimensional information can actually be effectively leveraged in DML without changing the base dimensionality and test time cost, which we motivate from two key elements. Firstly, in DML, an additional embedding space can be spanned by a multilayer perceptron (MLP) operating over the feature representation shared with the base embedding space (see e.g. \citet{milbich2020diva}). 
With larger dimensionalities, we can thus cheaply learn a secondary high-dimensional representation space simultaneously, also denoted as \textit{target} embedding space. 
Relative to the large feature backbone, and with the \emph{batchsize} capping the number of additional high dimensional operations, only little additional training cost is introduced. 
While we can not utilize the high-dimensional target embedding space at test time for aforementioned reasons, we may utilize it to boost the performance of the base embeddings.\\
Unfortunately, a simple connection of base and additional target embedding spaces through the shared feature backbone is insufficient for the base representation space to benefit from the auxiliary, high-dimensional information.
Thus, secondly, to efficiently leverage the high-dimensional context, we use insights from knowledge distillation \citep{hinton2015distilling}, where a small ``student'' model is trained to approximate a larger ``teacher'' model.
However, while knowledge distillation can be found in DML \citep{chen2017darkrank}, few-shot learning \citep{tian2020rethinking} and self-supervised extensions thereof \citep{rajasegaran2020selfsupervised}, the reliance on additional, commonly larger teacher networks or multiple training runs \citep{furlanello2018born}, introduces much higher training cost.
Fortunately, we find that the target embedding space learned \textit{simultaneously} at higher dimension can sufficiently serve as a ``teacher'' \textit{during} training - through knowledge distillation of its sample similarities, the performance of the base embedding space can be improved notably.
Such distillation intuitively encourages the lower-dimensional base embedding space to embed semantic similarities similar to the more expressive target embedding space and thus incorporate dimensionality-related generalization benefits.\\
Furthermore, \textit{S2SD} makes use of the low cost to span additional spaces to introduce multiple teacher spaces. Operating each of them at higher, but varying dimensionality, joint distillation can then be used to enforce reusability in the distilled content akin to feature reusability in meta-learning \citep{raghu2019rapid} for additional generalization boosts. 
Finally, in DML, the base embedding space is spanned over a penultimate feature space of much higher dimensionality, which introduces a dimensionality-based bottleneck \citep{milbich2020sharing}. By applying the distillation objective between feature and base embedding space in \textit{S2SD}, we further encourage better feature usage in base embedding space. This facilitates the approximation of high-dimensional context through the base embedding space for additional improvements in generalization.\\
The benefits to generalization are highlighted in performance boosts across three standard benchmarks, CUB200-2011 \citep{cub200-2011}, CARS196 \citep{cars196} and Stanford Online Products \citep{lifted}, where \textit{S2SD} improves test-set recall@1 of already strong DML objectives by up to $7\%$, while also setting a new state-of-the-art. Improvements are even more significant in very low dimensional base embedding spaces, making \textit{S2SD} attractive for large-scale retrieval problems which can benefit from reduced embedding dimensionalities.
Importantly, as \textit{S2SD} is applied \textbf{during} the same DML training process on the \textbf{same} network backbone, no large teacher networks or additional training runs are required. Simple experiments even show that \textit{S2SD} can outperform comparable 2-stage distillation at much lower cost.\\
In summary, our contributions can be described as:\\
\textbf{1)} We propose \textit{Simultaneous Similarity-based Self-Distillation} (\textit{S2SD}) for DML, using knowledge distillation of high-dimensional context without large additional teacher networks or training runs.\\
\textbf{2)} We motivate and evaluate this approach through detailed ablations and experiments, showing that the method is agnostic to choices in objectives, backbones, and datasets.\\
\textbf{3)} Across benchmarks, we achieve significant improvements over strong baseline objectives and state-of-the-art performance, with especially large boosts for very low-dimensional embedding spaces.

\section{Related Work}
\input{figures/setup}
\textbf{Deep Metric Learning (DML)} has proven useful for zero-shot image/video retrieval \& clustering \citep{semihard,margin,Brattoli_2020_CVPR}, face verification \citep{sphereface,arcface} and contrastive (self-supervised) representation learning (e.g. \citet{moco,chen2020simple,pretextmisra}).
Approaches can be divided into \textbf{1)} improved ranking losses, \textbf{2)} tuple sampling methods and \textbf{3)} extensions to the standard DML training approach.
\textbf{1)} Ranking losses place constraints on the relations available in image tuples, ranging from pairs (s.a. \citet{contrastive}) to triplets \citep{semihard} and more complex orderings \citep{quadtruplet,lifted,npairs,multisimilarity}.
\textbf{2)} As the number of possible tuples scales exponentially with dataset size, tuple sampling approaches have been introduced to tackle tuple redudancy and to ensure that meaningful tuples are presented during training. These tuple sampling methods can follow heuristics \cite{semihard,margin}, be of hierarchical nature \citep{htl} or learned \citep{roth2020pads}. 
Similarly, learnable proxies to replace tuple members \cite{proxynca,kim2020proxy,softriple} can also remedy the sampling issue, which can be extended to tackle DML from a classification viewpoint \citep{zhai2018classification,arcface}.
\textbf{3)} Finally, extensions to the basic training scheme can involve synthetic data \citep{dvml,hardness-aware,daml}, complementary features \citep{mic,milbich2020diva}, a division into subspaces \citep{Sanakoyeu_2019_CVPR,dreml,abe,abier}, training of multiple networks \citep{diversifieddml} using mutual learning \cite{mutual_learning} or higher-order moments \citep{horde}. 
\textit{S2SD} can similarly be seen as an extension to DML, though we specifically focus on capturing and distilling complex high-dimensional sample relations within lower dimensional embedding spaces to improve generalization.\\
\textbf{Knowledge Distillation} involves knowledge transfer from teacher to (usually smaller) student models, e.g. by matching network softmax outputs/logits \citep{bucilu2006model,hinton2015distilling}, (attention-weighted) feature maps \citep{romero2014fitnets,zagoruyko2016paying}, or latent representations \citep{Ahn_2019,Park_2019,tian2019contrastive,laskar2020dataefficient}.
Importantly, \citet{tian2019contrastive} show that under fair comparison, basic matching via Kullback-Leibler (KL) Divergences as used in \citet{hinton2015distilling} performs very well, which we also find to be the case for \textit{S2SD}. 
This is further supported in recent few-shot learning literature \citep{tian2020rethinking}, wherein KL-distillation alongside self-distillation (by iteratively reusing the same network as a teacher for beneficial generalization and regulatory properties \citep{furlanello2018born,selfreference,self_distill_1,self_distill_2,self_distill_3}) in additional meta-training stages improves feature representation strength important for generalization \citep{raghu2019rapid}.

Our work is closest to \citet{beyourownteacher} and \citet{MetaDistiller}, which propose to break down a network into a cascading set of subnetworks, wherein each subsequent subnetwork builds on its predecessors. In doing so, each subnetwork is trained independently on a classification task at hand. Knowledge distillation is then applied either from the full network \citep{beyourownteacher} acting as a teacher or via soft targets generated from a meta-learned label generator \citep{MetaDistiller}, to each smaller student subnetwork during the same training run to improve overall performance.
In a related manner, \textit{S2SD} utilizes similar concurrent, but relational self-distillation to instead encode high-dimensional sample relation context from multiple, higher-dimensional teacher embedding spaces; this is crucial to improve the generalization capabilities of a single student embedding space for zero-shot, out-of-distribution image retrieval tasks. As such, it operates orthogonally to proposals made by \citet{beyourownteacher} and \citet{MetaDistiller}. The concurrency of the self-distillation in turn is a consequence of the novel insight that solely the dimensionality of embedding spaces can serve as meaningful teachers, as these can be spanned cheaply over a large, shared feature backbone.\\ 
The novel dimensionality-based concurrent distillation also sets S2SD apart from existing knowledge distillation applications to DML, which are done in a generic manner with separate, larger teacher networks or additional training stages \citep{chen2017darkrank,Yu_2019,Han2019DeepDM,laskar2020dataefficient}.


\section{Method}\label{sec:methods}
We now introduce key elements for \textit{Simultaneous Similarity-based Self-Distillation }\textit{(S2SD)} to improve generalization of embedding spaces by utilizing higher dimensional context. We begin with the preliminary notation and fundamentals to Deep Metric Learning (\S\ref{sec:prelim}), before defining the three key elements to \textit{S2SD}: Firstly, the Dual Self-Distillation (DSD) objective, which uses KL-Distillation on a concurrently learned embedding space of higher dimensionality (\S\ref{sec:dual}) to introduce important high-dimensional context into a low-dimensional embedding space during training. We then extend this to Multiscale Self-Distillation (MSD) with distillation from several different high-dimensional embedding spaces to encourage reusability in the distilled context (\S\ref{sec:multi}). Finally, we shift to self-distillation from normalized feature representations (MSDF) to counter dimensionality bottlenecks commonly encountered in DML (\S\ref{sec:feats}).

\subsection{Preliminaries}
\label{sec:prelim}
DML builds on generic Metric Learning which aims to find a (parametrized) distance metric $d_\theta: \Phi\times \Phi \mapsto \mathbb{R}$ on the \textit{feature space} $\Phi\subset\mathbb{R}^{d^*}$ over images $\mathcal{X}$ that best satisfy ranking constraints usually defined over class labels $\mathcal{Y}$. This holds also for DML. However, while Metric Learning relies on a \textbf{fixed} feature extraction method to obtain $\Phi$, DML introduces deep neural networks to concurrently learn a feature representation. 
Most such DML approaches aim to learn Mahalanobis distance metrics, which cover the parametrized family of inner product metrics \citep{surez2018tutorial,chen2019curvilinear}. 
These metrics, with some restrictions \citep{surez2018tutorial}, can be reformulated as
\begin{align}\label{eq:eucl_dist}
\begin{split}
    d(\phi_1,\phi_2) &= \sqrt{(L(\phi_1-\phi_2)^TL(\phi_1-\phi_2)} \\
                     &= \left\Vert L\phi_1-L\phi_2\right\Vert_2 = \left\Vert \psi_1-\psi_2\right\Vert_2
\end{split}
\end{align}
with learned linear projection $L\in\mathbb{R}^{d\times d^*}$ from $d^*$-dim. \textit{features} $\phi_i\in\Phi$ to $d$-dim. \textit{embeddings} $\psi_i:=(f\circ\phi)(x_i)\in\Psi_f$ with embedding function $f:\phi_i\mapsto L\phi_i$. Importantly, this redefines the motivation behind DML as learning $d$-dimensional image embeddings $\psi$ s.t. their euclidean distance $d(\bullet,\bullet)=\left\Vert\bullet-\bullet\right\Vert_2$ is connected to semantic similarities in $\mathcal{X}$. This embedding-based formulation offers the significant advantage of being compatible with fast approximate similarity search methods (e.g. \cite{faiss}), allowing for large-scale applications at test time.
In this work, we assume $\Psi_f$ to be normalized to the unit hypersphere $\mathcal{S}_{\Psi_f}$, which is commonly done \citep{margin,Sanakoyeu_2019_CVPR,sphereface,wang2020understanding} for beneficial regularizing purposes \citep{margin,wang2020understanding}. 
For the remainder we hence set $\Psi$ to refer to $S_{\Psi}$.\\
Common approaches to learn such a representation space involve training surrogates on ranking constraints defined by class labels. Such approaches start from pair or triplet-based ranking objectives \citep{contrastive,semihard}, where the latter is defined as
\begin{align}
    \mathcal{L}_\text{triplet} &= \frac{1}{|\mathcal{T}_\mathcal{B}|}\textstyle{\sum}_{(x_i,x_j,x_k)\in\mathcal{T}_\mathcal{B}} \left[d(\psi_i,\psi_j) - d(\psi_i,\psi_k)+ m\right]_+
\end{align}
with margin $m$ and the set of available triplets $(x_i,x_j,x_k)\in\mathcal{T}_\mathcal{B}$ in a mini-batch $\mathcal{B}\subset\mathcal{X}$, with $y_i=y_j\neq y_k$.
This can be extended with more complex ranking constraints or tuple sampling methods. 
We refer to Supp. \ref{supp:base_methods} and \citet{roth2020revisiting} for further insights and detailed studies.

\subsection{Embedding Space Self-Distillation}\label{sec:dual}
For the aforementioned standard DML setting, generalization performance of a learned embedding space can be linked to the utilized embedding dimensionality. However, high dimensionality results in notably higher retrieval cost on downstream applications, which limits realistically usable dimensions.
In \textit{S2SD}, we show that high-dimensional context can be used as a teacher during the training run of the low-dimensional \textit{base} or \textit{reference} embedding space. As the base embedding model is also the one that is evaluated, test time retrieval costs are left unchanged.

To achieve this, we simultaneously train an additional high-dimensional \textit{auxiliary}/\textit{target} embedding space $\Psi_{g}:=(g\circ\phi)(\mathcal{X})$ spanned by a secondary embedding branch $g$. $g$ is parametrized by a MLP or a linear projection, similar to the base embedding space $\Psi_f$ spanned by $f$, see \S\ref{sec:prelim}. Both $f$ and $g$ operate on the same large, shared feature backbone $\phi$.
For simplicity, we train $\Psi_f$ and $\Psi_{g}$ using the same DML objective $\mathcal{L}_\text{DML}$.

Unfortunately, higher expressivity and improved generalization of high-dimensional embeddings in $\Psi_g$ hardly benefit the base embedding space, even with a shared feature backbone. To explicitly leverage high-dimensional context for our base embedding space, we utilize knowledge distillation from target to base space.
However, while common knowledge distillation approaches match single embeddings or features between student and teacher, the different dimensionalities in $\Psi_f$ and $\Psi_g$ inhibit naive matching. 

Instead, \textit{S2SD} matches sample relations (see e.g. \cite{tian2019contrastive}) defined over batch-similarity matrices $D\in\mathbb{R}^{\mathcal{B}\times\mathcal{B}}$ in base and target space, $D^f$ and $D^{g}$, with batchsize $\mathcal{B}$. We thus encourage the base embedding space to relate different samples in a similar manner to the target space. 
To compute $D$, we use a cosine similarity by default, given as $D_{i,j}=\psi_i^T\psi_j$, since $\psi_i$ is normalized to the unit hypersphere. Defining $\sigma_\text{max}$ as the softmax operation and $\mathcal{D}_\text{KL}(p,q) = \sum \log(p)\nicefrac{\log(p)}{\log(q)}$ as the Kullback-Leibler-divergence, we thus define the simultaneous self-distillation objective as
\begin{equation}\label{eq:kl_distill}
    \mathcal{L}_\text{dist}(D^{f}, D^{g}) = \textstyle\sum_i^{|\mathcal{B}|} \mathcal{D}_\text{KL}\left(\sigma_\text{max}\left(\nicefrac{D^f_{i,:}}{T}\right), \sigma_\text{max}^\dagger\left(\nicefrac{D^{g}_{i,:}}{T}\right)\right)
\end{equation}
with temperature $T$, as visualized in Figure \ref{fig:setup}. ($^\dagger$) denotes no gradient flow to target branches $g$ as we only want the base space to learn from the target space.
By default, we match rows or columns of $D$, $D_{i,:}$, effectively distilling the relation of an anchor embedding $\psi_i$ to all other batch samples. 
Embedding all batch samples in base dimension, $\Psi_f^\mathcal{B}: \mathcal{B}\mapsto \psi_f(\mathcal{B})$, and higher dimension, $\Psi_{g}^\mathcal{B}: \mathcal{B}\mapsto \psi_{g}(\mathcal{B})$, the (simultaneous) \emph{Dual Self-Distillation} (DSD) training objective then becomes
\begin{align}\label{eq:dual}
\begin{split}
    \mathcal{L_\text{DSD}}(\Psi_f^\mathcal{B},\Psi_{g}^\mathcal{B}) &= \nicefrac{1}{2}\cdot\left[\mathcal{L}_\text{DML}(\Psi_f^\mathcal{B}) + \mathcal{L}_\text{DML}(\Psi_{g}^\mathcal{B})\right] \\
    &+ \gamma\cdot \mathcal{L}_\text{dist}(D^f,D^{g})
\end{split}
\end{align}


\subsection{Reusable Relations by Multiscale Distillation}\label{sec:multi}
While \textit{DSD} encourages the reference embedding space to recover complex sample relations by distilling from a higher-dimensional target space spanned by $g$, it is not known \textit{a priori} which distillable sample relations actually benefit generalization of the reference space.\\
To encourage the usage of sample relations that more likely aid generalization, we follow insights made in \citet{raghu2019rapid} on the connection between \textbf{reusability} of features across multiple tasks and better generalization thereof. We motivate reusability in \textit{S2SD} by extending \textit{DSD} to \emph{Multiscale Self-Distillation} (\textit{MSD}) with distillation instead from $m$ multiple different target spaces spanned by $G=\{g_k\}_{k\in\{1,m\}}$. Importantly, each of these high-dimensional target spaces operate on different dimensionalities, i.e. $dim\text{ } f < dim\text{ } g_1 < ... < dim\text{ } g_{m-1} < dim\text{ } g_m$. As this results in each target embedding space encoding sample relations differently, application of distillation across all spaces spanned by $G$ pushes the base branch towards learning from sample relations that are reusable across all higher dimensional embedding spaces and thereby more likely to generalize (see also Fig. \ref{fig:setup}).

Specifically, given the set of target similarity matrices $\{D^k\}_{k\in\{f,g_1,...,g_m\}}$ and target batch embeddings $\Gamma^{m}:=\{\Psi_k^\mathcal{B}\}_{k\in\{f,g_1,...,g_m\}}$, we then define the \textit{MSD} training objective as
\begin{align}\label{eq:multi}
\begin{split}
    \mathcal{L_\text{MSD}}(\Gamma^{m}) &= \frac{1}{2}\cdot\left[\mathcal{L}_\text{DML}(\Psi_f^\mathcal{B}) + \frac{1}{m}\textstyle\sum_{i=1}^m\mathcal{L}_\text{DML}(\Psi_{g_i}^\mathcal{B})\right] \\
    &+ \frac{\gamma}{m}\textstyle\sum_{i=1}^m \mathcal{L}_\text{dist}(D^f,D^{g_i})    
\end{split}
\end{align}

\subsection{Tackling the Dimensionality Bottleneck by Feature Space Self-Distillation}\label{sec:feats}
As noted in \S\ref{sec:prelim}, the base embedding $\Psi$ utilizes linear projections $f$ from the (penultimate) feature space $\Phi$ where $dim\text{ } \Phi$ is commonly much larger than $dim\text{ } \Psi$. While compressed semantic spaces encourage stronger representations \citep{alex2016deep,dai2019diagnosing} to be learned, \citet{milbich2020sharing} show that the actual test performance of the lower-dimensional embedding space $\Phi$ is inferior to that of the non-adapted, but higher-dimensional feature space $\Psi$.

This supports a dimensionality-based loss of information beneficial to generalization, which can hinder the base embedding space to optimally approximate the high-dimensional context introduced in \S\ref{sec:dual} and \ref{sec:multi}.\\
To rectify this, we apply self-distillation following eq. \ref{eq:kl_distill} on the normalized feature representations $\Phi^n$ generated by normalizing the backbone output $\phi$. 
With the batch of normalized feature representations $\Psi_{\phi^n}^\mathcal{B}$ we get \emph{multiscale self-distillation with feature distillation} (\textit{MSDF}) (see also Fig.  \ref{fig:setup})
\begin{equation}\label{eq:feat_comp}
    \mathcal{L}_\text{MSDF}(\Gamma^m,\Psi_{\phi^n}^\mathcal{B}) = \mathcal{L}_\text{MSD}(\Gamma^m) + \gamma\mathcal{L}_\text{dist}(D^f,D^{\phi^n})
\end{equation}
In the same manner, one can also address other architectural information bottlenecks such as through the generation of feature representations from a single global pooling operation. While not noted in the original publication, \citet{kim2020proxy} address this in the official code release by using both global max- and average pooling to create their base embedding space. While this naive usage changes the architecture at test time, in \textit{S2SD} we can \textit{fairly} leverage potential benefits by \textit{only} spanning the auxiliary spaces (and distilling) from such feature representations (denoted as \textit{DSDA}/\textit{MSDA}/\textit{MSDFA}).

\section{Experimental Setup}\label{sec:experiments}
\input{tables/Short_Full_Results}

\input{tables/sota}
We study \textit{S2SD} in four experiments to establish 1) method ablation performance \& relative improvements, 2) state-of-the-art, 3) comparisons to standard 2-stage distillation, benefits to low-dimensional embedding spaces \& generalization properties and 4) motivation for architectural choices.

\textbf{Method Notation.} We abbreviate ablations of \textit{S2SD} (see \S\ref{sec:methods}) in our experiments as: 
\textit{DSD} \& \textit{MSD} for \textbf{D}\textit{ual} (\ref{sec:dual}) \& \textbf{M}\textit{ultiscale} \textbf{S}\textit{elf}-\textbf{D}\textit{istillation} (\ref{sec:multi}), \textit{MSDF} the addition of \textbf{F}\textit{eature distillation} (\ref{sec:feats}) and \textit{DSDA}/\textit{MSD(F)A} the inclusion of multiple pooling operations in the auxiliary branches (also \S\ref{sec:feats}).

\subsection{Experiments}
\textbf{Fair Evaluation of \textit{S2SD}.} \S\ref{sec:baselines} specifically applies \textit{S2SD} and its ablations to three DML baselines. To show realistic benefit, \textit{S2SD} is applied to best-performing objectives evaluated in \citet{roth2020revisiting}, namely \textit{(i)} Margin loss with Distance-based Sampling \citep{margin}, \textit{(ii)} their proposed Regularized Margin loss and \textit{(iii)} Multisimilarity loss \citep{multisimilarity}, following their experimental training pipeline. This setup utilizes no learning rate scheduling and fixes common implementational factors of variation in DML pipelines such as batchsize, base embedding dimension, weight decay or feature backbone architectures to ensure comparability in DML (more details in Supp. \ref{supp:impl_dets}). 
As such, our results are directly comparable to their large set of examined methods and guaranteed that relative improvements solely stem from the application of \textit{S2SD}.

\textbf{Comparison to literature.} \S\ref{sec:comp} further highlights the benefits of \textit{S2SD} by comparing \textit{S2SD}'s boosting properties across literature standards, with different backbone architectures and base embedding dimensions: \textit{(1)} ResNet50 with $d$ = 128 \citep{margin,mic} and \textit{(2)} $d$ = 512 \citep{zhai2018classification} as well as \textit{(3)} variants to Inception-V1 with Batch-Normalization at $d$ = 512 \citep{multisimilarity,softriple,milbich2020diva}.
Only here do we conservatively apply learning rate scheduling, since all references noted in Table \ref{tab:sota} employ scheduling as well. We categorize published work based on backbone architecture and embedding dimension for fairer comparison. Note that this is a less robust comparison than done in \S \ref{sec:baselines}, due to potential implementation differences between our pipeline and reported literature results.

\textbf{Comparison to 2-Stage Distillation.} \S\ref{sec:further_exp} compares \textit{S2SD} to 2-stage distillation, investigates benefits to very low dimensional reference spaces and examines the connection between improvements and changes in embedding space density and spectral decay (see Supp. \ref{supp:gen_metrics}), which have been linked to improved generalization.%

\textbf{Ablation Study.} \S\ref{sec:ablations} ablates and motivates architectural choices in \textit{S2SD} used throughout \S \ref{sec:experiments}. Pseudo code and detailed results are available in Supp. \ref{supp:pseudo_code}, \ref{supp:detailed_1}, and \ref{supp:detailed_2}.

\subsection{Implementation}
\textbf{Datasets \& Evaluation.}
In all experiments, we evaluate on standard DML benchmarks: \textit{CUB200-2011} \citep{cub200-2011}, \textit{CARS196} \citep{cars196} and \textit{Stanford Online Products (SOP)} \citep{lifted}.
Performance is measured in \textit{recall at 1} (R@1) and \textit{at 2} (R@2) \citep{recall} as well as \textit{Normalized Mutual Information} (NMI) \citep{nmi}. Results measured on mean Average Precision evaluated on Recall (mAP@R) are available in the Supplementary along with additional dataset details.

\textbf{Experimental Details.}
Our implementation follows \citet{roth2020revisiting} for comparability, using frozen Batch-Normalization \cite{batchnorm} whenever a ResNet50 \cite{resnet} is utilized, Adam \cite{adam} with learning rate $10^{-5}$ for training and weight decay \cite{weight_decay} of $4\cdot 10^{-5}$ for regularization. Additional details are available in Supp. (\ref{supp:bench_dets}).
For \S\ref{sec:baselines}-\ref{sec:ablations}, we only adjust the respective pipeline elements in questions.
For \textit{S2SD}, unless noted otherwise (s.a. in \S \ref{sec:ablations}), we set $\gamma=50, T=1$ for all objectives on CUB200 and CARS196, and $\gamma=5, T=1$ on SOP. \textit{DSD} uses target-dim. $d=2048$ and MSD uses target-dims. $d\in[512,1024,1536,2048]$.
We found it beneficial to activate the feature distillation after $n=1000$ iterations to ensure that meaningful features are learned first before feature distillation is applied. Additional embedding spaces are spanned by two layer MLPs with row-wise KL-distillation of high-dimensional similarities (eq. \ref{eq:kl_distill}), applied as in $\mathcal{L}_\text{multi}$ (eq. \ref{eq:multi}). By default, we use Multisimilarity Loss as stand-in for $\mathcal{L}_\text{DML}$. Hyperparameters were determined previous to the result runs using a 80-20 training and validation split, similar to \citet{roth2020revisiting} and \citet{kim2020proxy}.

\section{Results}
\subsection{Fair performance study} \label{sec:baselines}
In Tab. \ref{tab:relative_results} (full table in Supp. Tab. \ref{tab:relative_results_long}), we show that under fair experimental protocol, utilizing \textit{S2SD} and its ablations gives an objective and benchmark independent, significant boost in performance by up to $7\%$ opposing the existing DML objective performance plateau. This holds even for previous state-of-the-art regularized objectives s.a. R-Margin loss as well as proxy-based objectives such as ProxyAnchor (\citet{kim2020proxy}, see Supplementary), highlighting the effectiveness of \textit{S2SD} for DML. Across objectives, \textit{S2SD}-based changes in wall-time do not exceed negligible $5\%$ with only minor convergence impacts. 
\input{figures/metrics}
\subsection{Setting a new State-of-the-Art} 
\label{sec:comp}
Motivated by Tab. \ref{tab:relative_results}, we use \textit{MSDFA} for CUB200/CARS196 and \textit{MSDF} for SOP. Table \ref{tab:sota} shows that \textit{S2SD} can boost baseline objectives to reach and even surpass other state-of-the-art methods, in parts with a notable margin. This holds even when compared to much more complex methods with feature mining or RL-policies such as MIC \citep{mic}, DiVA \citep{milbich2020diva} or PADS \citep{roth2020pads}, to which \textit{S2SD} operates orthogonally. Finally, we note that these insights are true even with our results reported with confidence intervals, which is commonly neglected in DML. 

\input{figures/ablations_no1}
\subsection{Benefits of S2SD}\label{sec:further_exp}
%
\textbf{Comparison to standard distillation.} 
With a student \textit{S} using the same objective and embedding 
dimensionality as the reference branch in \textit{DSD})
and a teacher \textit{T} at the highest optimal dimensionality $d=2048$, 
we find that separating \textit{DSD} into a standard 2-stage distillation setup actually degenerates performance (see Fig. \ref{fig:ablations_1}A, compare to \textit{Dist.}). 
In addition, \textit{S2SD} allows for easy integration of teacher ensembles, realized by \textit{MSD(F,A)}, to even outperform the teacher by a notable margin. This is specifically interesting as \textit{S2SD} retains the operating embedding dimensionality of the student.

\textbf{Benefits to lower base dimensions.} We now show that our module is able to vastly boost networks limited to very low embedding dimensions, which we visualize in figure \ref{fig:ablations_1}B). 
For example, networks operating on $d=32\text{ \& }64$ trained with \textit{S2SD} can match the performance of networks trained and evaluated on embedding dimensions \textit{four or eight times} the size. For $d=128$, \textit{S2SD} even outperforms the highest dimensional baseline at $d=2048$ by a large margin.

\textbf{Embedding space metrics.} 
We now look at relative changes in embedding space density
and spectral decay (see supplementary of \citet{roth2020revisiting}) when applying \textit{S2SD}.
Our study, visualized in figure \ref{fig:metrics}, shows that the application of \textit{S2SD} increases embedding space density and lowers the spectral decay (thus providing a more feature-diverse embedding space) across criteria, which is aligned with properties of improved generalization in DML as noted in \citet{roth2020revisiting}. 

\subsection{Motivating S2SD Architecture Choices}\label{sec:ablations}
\textbf{Distillation improves generalization through \textit{S2SD}.} Fig. \ref{fig:ablations_1}A (\textit{Joint}) and Fig. \ref{fig:ablations_1}F ($\gamma=0$) highlight how crucial self-distillation is, as incorporating a secondary embedding space without any distillation link hardly improves performance. 
Fig. \ref{fig:ablations_1}A (\textit{Concur.}) further shows that joint training of a detached reference embedding $f$, while otherwise training in high dimension, similarly doesn't offer notable improvement. 
Finally, Figure \ref{fig:ablations_1}F shows robustness to changes in $\gamma$, with peaks around $\gamma=50$ and $\gamma=5$ for CUB200/CARS196 and SOP. We also found best performance for temperatures $T\in[0.2,2]$ and hence set $T=1$ by default.

\textbf{Best way to enforce reusability.} To motivate our many-to-one self-distillation $\mathcal{L}_\text{MSD}$ (eq. \ref{eq:multi}, here also dubbed $\mathcal{L}_\text{Multi}$), we evaluate against other distillation setups that could support reusability of distilled sample relations: 
\textit{(1)} \textit{Nested} distillation, where instead of distilling all target spaces only to the reference space, we distill from a target space to \textit{all} lower-dimensional embedding spaces:
\begin{align}\label{eq:multi_nested}
\begin{split}
    \mathcal{L_\text{Nested}}&(\Gamma^{m}) = \frac{1}{2}\left[\mathcal{L}_\text{DML}(\Psi_f^\mathcal{B}) + \frac{1}{m}\sum_{i=1}^m\mathcal{L}_\text{DML}(\Psi_{g_i}^\mathcal{B})\right]\\
    &+\frac{\gamma}{{m\choose m-1}}\sum_{\substack{i=0,j=1, j\neq i\\dim\text{ }g_j\geq dim\text{ }g_i}}^{m} \mathcal{L}_\text{dist}(\Psi_{g_i}^\mathcal{B},\Psi_{g_j}^\mathcal{B})
\end{split}
\end{align}
In the second term, $g_0$ denotes the base embedding $f$.\\ 
\textit{(2)} \textit{Chained} distillation, which distills target spaces only to the immediate lower-dimensional embedding space:
\begin{align}\label{eq:multi_chained}
\begin{split}
    \mathcal{L_\text{Chained}}(\Gamma^m) &= \frac{1}{2}\left[\mathcal{L}_\text{DML}(\Psi_f^\mathcal{B}) + \frac{1}{m}\sum_{i=1}^m\mathcal{L}_\text{DML}(\Psi_{g_i}^\mathcal{B})\right]\\
    &+\frac{\gamma}{m}\sum_{i=1}^{m-1} \mathcal{L}_\text{dist}(\Psi_{g_i}^\mathcal{B},\Psi_{g_{i-1}}^\mathcal{B})
\end{split}
\end{align}
Figure \ref{fig:ablations_1}E shows that a many-to-one distillation performs notably better, supporting the reusability aspect and $\mathcal{L}_\text{multi}$ as our default method.

\textbf{Choice of distillation method \& branch structures.} Fig.  \ref{fig:ablations_1}C evaluates various distillation objectives, finding KL-divergence between vectors of similarities to perform better than KL-divergence applied over full similarity matrices or row-wise means thereof, as well as cosine/euclidean distance-based distillation (see e.g. \citet{Yu_2019}). 
Figure \ref{fig:ablations_1}D shows insights into optimal auxiliary branch structures, with two-layer MLPs giving the largest benefit, although even a linear target mapping reliably boosts performance. This coincides with insights made by \citet{chen2020simple}. Further network depth only deteriorates performance.

\section{Conclusion}
In this paper, we propose a novel DML training paradigm based on dimensionality-based knowledge distillation, \textit{Simultaneous Similarity-based Self-Distillation} (\textit{S2SD)}. \textit{S2SD} allows for the inclusion of reusable, context-rich, high-dimensional relational information for improved generalization. 
This is achieved by solving the standard DML objective simultaneously in higher-dimensional embedding spaces while applying knowledge distillation concurrently between these high-dimensional teacher spaces and a lower-dimensional reference space. In doing so, \textit{S2SD} introduces little additional computational overhead, with no extra cost at test time. Thorough ablations and experiments show \textit{S2SD} significantly improving the generalization performance of existing DML objectives regardless of embedding dimensionality, thereby also setting a new state-of-the-art.

\newpage
\section*{Acknowledgements}
We would like to thank Samarth Sinha (University of Toronto, Vector), Matthew McDermott (MIT) and Mengye Ren (University of Toronto, Vector) for insightful discussions and feedback on the paper draft. This work was funded in part by a CIFAR AI Chair at the Vector Institute, Microsoft Research, and an NSERC Discovery Grant. Resources used in preparing this research were provided, in part, by the Province of Ontario, the Government of Canada through CIFAR, and companies sponsoring the Vector Institute \url{www.vectorinstitute.ai/#partners}.

\section*{Reviewer Comments}
\textbf{Re: More advanced baseline methods.}
We do believe that our reported results are representative of the current state of Deep Metric Learning - under fair comparison \citep{roth2020revisiting}, Multisimilarity and Margin loss achieve best or competitive results. In addition, \textit{S2SD} is applied to regularized Deep Metric Learning (R-Margin loss \citep{roth2020revisiting}) and shows high improvements throughout.
ProxyAnchor \citep{kim2020proxy} was not included in the literature comparison due to different architecture settings (lines 268-271), thus not allowing for a fair comparison. DiVA in turn offers a separate extension to DML and should be seen as orthogonal to \textit{S2SD}, which however we consistently outperform. 
However, for completeness, we have applied \textit{S2SD} to ProxyAnchor using the fair comparison setup in \citet{roth2020revisiting} (see section 5.1) on all benchmarks to highlight the general applicability of \textit{S2SD}. We perform no hyperparameter tuning and use those already mentioned in the paper and find notable and consistent performance improvements. Results are available in the supplementary (Table \ref{tab:proxyanchor_results}).

\textbf{Re: Runtime Analysis.} 
We offer a dimensionality versus runtime analysis in the supplementary (Table \ref{tab:test_time}), which shows significant runtime reduction by up to a magnitude: With performance of $d=64$ roughly matching that of $d=2048$, retrieval runtime on $N=250000$ synthetic evaluation samples can be reduced from 27.21 $\pm$ 0.17 to 1.98 $\pm$ 0.00 seconds. The high performance boost also offers the benefit of reduced memory storage needed to retain embedding vectors for comparable retrieval performance.

\textbf{Re: Necessity of loss terms.}
There are only three effective loss terms. Firstly, the default DML objective $\mathcal{L}_\text{DML}$ applied to each branch.
Secondly, the actual distillation objective over all branches $\frac{\gamma}{m}\sum_{i=1}^m \mathcal{L}_\text{Dist}(D^f, D^{g_i})$ (see eq. \ref{eq:multi}) and thirdly, the feature space self-distillation to counter the dimensionality bottleneck (see eq. \ref{eq:feat_comp}). These last two correspond to the S2SD variants MSD and MSDF, which we evaluate against different variants of $\mathcal{L}_\text{DML}$ in section \ref{sec:baselines}. 
As can be seen, both loss-terms operate complementary. Regarding general convergence behaviour and training dynamics, we find only minor changes, with maximum performance achieved at similar training stages. 

\textbf{Re: Validation Information.}
All benchmarks only offer a train/test split. As such, we use a 80-20 train/validation split of the original training split to determine hyperparameters (e.g. \citet{roth2020revisiting} and \citet{kim2020proxy}), and use those for training on the full training dataset and evaluation on the test set used throughout literature (see Sec. \ref{sec:experiments}). 

\textbf{Re: Theoretical Grounding.}
While we do not offer detailed theoretical formalism in this paper as to why \textit{S2SD} is so effective, we do evaluate changes in the embedding space structure (see figure 2), which highlight that \textit{S2SD} encourages noticeably higher feature diversity and embedding space uniformity linked to improved generalization \citep{roth2020revisiting,wang2020understanding,milbich2020diva}.

\bibliography{icml2021}
\bibliographystyle{icml2021}

\newpage
\appendix
\input{appendix}

\end{document}

%% file: figures/setup.tex
\begin{figure*}[t]
    \centering
    \includegraphics[width=1\textwidth]{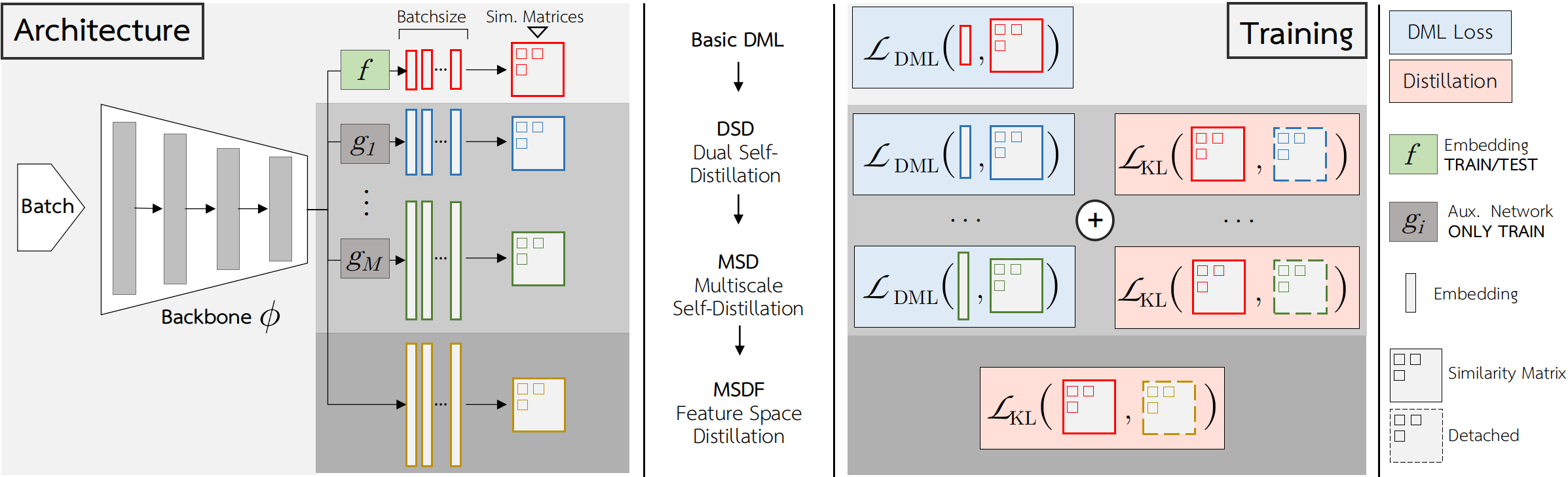}
    \caption{\textit{S2SD.} We use a standard encoder $\phi$, embedding $f$, and multiple auxiliary embedding networks $g_i$ (used only during training) depending on the \textit{S2SD} approach used. 
    During training, for each batch of embeddings produced by the respective embedding network $g_i$, we compute DML losses while applying embedding distillation on the respective batch-similarity matrices (\textit{DSD/MSD}). We further distill from the feature representation space for additional information gain (\textit{MSDF}).}
    \label{fig:setup}
\end{figure*}

%% file: tables/Short_Full_Results.tex
\begin{table*}[t]
    \caption{\textit{S2SD comparison against strong baseline objectives.} \textbf{Bold} denotes best results per objective, \blue{\textbf{bluebold}} marks best overall results. mAP@R results as proposed in \cite{roth2020revisiting} and \cite{musgrave2020metric} as well as ProxyAnchor evaluations (\citet{kim2020proxy}, using a different setup) can be found in the Supplementary (Table \ref{tab:relative_results_map} and \ref{tab:proxyanchor_results}), further showing the notable benefits of S2SD.}
 \footnotesize
  \setlength\tabcolsep{1.4pt}
  \centering
  \resizebox{\textwidth}{!}{
  \begin{tabular}{l || c | c || c | c || c | c}
     \toprule
     \multicolumn{1}{l}{\textsc{Benchmarks}$\rightarrow$} & \multicolumn{2}{c}{\textsc{CUB200-2011}} & \multicolumn{2}{c}{\textsc{CARS196}} & \multicolumn{2}{c}{\textsc{SOP}} \\
     \midrule
     \textsc{Approaches} $\downarrow$ & R@1 & NMI & R@1 & NMI & R@1 & NMI\\
    \midrule
    \rowcolor{vvlightgray}
    \textbf{Margin}, $\beta=1.2$, \citep{margin} & $63.09\pm0.46$ & $68.21\pm0.33$ & $79.86\pm0.33$ & $67.36\pm0.34$ & $78.43\pm0.07$ & $90.40\pm0.03$\\        
    + DSD & $65.11\pm0.18$ & $69.65\pm0.44$ & $83.19\pm0.18$ & $69.28\pm0.56$ & $79.05\pm0.12$ & $90.52\pm0.18$\\
    + MSD & $66.13\pm0.34$ & $70.83\pm0.27$ & $83.63\pm0.31$ & $69.80\pm0.36$ & $79.26\pm0.15$ & $90.60\pm0.10$\\
    + MSDF & $\mathbf{67.58\pm0.32}$ & $\mathbf{71.47\pm0.19}$ & $85.55\pm0.23$ & $\mathbf{71.68\pm0.54}$ & \blue{$\mathbf{79.63\pm0.15}$} & \blue{$\mathbf{90.70\pm0.09}$} \\
    + MSDFA & $67.21\pm0.23$ & $71.43\pm0.25$ & $\mathbf{86.45\pm0.35}$ & $71.46\pm0.24$ & $78.82\pm0.09$ & $90.49\pm0.06$ \\
    \midrule
    \rowcolor{vvlightgray}
    \textbf{R-Margin}, $\beta=0.6$, \citep{roth2020revisiting} & $64.93\pm0.42$ & $68.36\pm0.32$ & $82.37\pm0.13$ & $68.66\pm0.47$ & $77.58\pm0.11$ & $90.42\pm0.03$\\ 
    + DSD & $66.58\pm0.08$ & $70.03\pm0.41$ & $84.64\pm0.16$ & $70.87\pm0.18$ & $77.86\pm0.10$ & $90.50\pm0.03$\\
    + MSD & $66.81\pm0.27$ & $70.47\pm0.16$ & $85.01\pm0.10$ & $71.67\pm0.40$ & $78.00\pm0.06$ & $90.47\pm0.04$\\
    + MSDF & $68.12\pm0.30$ & $\mathbf{71.80\pm0.33}$ & $85.78\pm0.22$ & $\mathbf{72.24\pm0.31}$ & $\mathbf{78.57\pm0.09}$ & $\mathbf{90.58\pm0.02}$\\
    + MSDFA & $\blue{\mathbf{68.58\pm0.26}}$ & $71.64\pm0.40$ & $\blue{\mathbf{86.81\pm0.35}}$ & $71.48\pm0.29$ & $78.00\pm0.11$ & $90.41\pm0.02$\\
    \midrule
    \rowcolor{vvlightgray}
    \textbf{Multisimilarity} \citep{multisimilarity} & $62.80\pm0.70$ & $68.55\pm0.38$ & $81.68\pm0.19$ & $69.43\pm0.38$ & $77.99\pm0.09$ & $90.00\pm0.02$\\
    + DSD & $65.57\pm0.26$ & $70.08\pm0.33$ & $83.51\pm0.20$ & $70.30\pm0.05$ & $78.23\pm0.04$ & $90.08\pm0.04$\\
    + MSD & $65.80\pm0.16$ & $70.66\pm0.01$ & $83.98\pm0.10$ & $71.34\pm0.09$ & $78.42\pm0.09$ & $90.09\pm0.03$\\        
    + MSDF & $67.04\pm0.29$ & $\blue{\mathbf{71.87\pm0.19}}$ & $85.69\pm0.19$ & $\blue{\mathbf{72.77\pm0.13}}$ & $\mathbf{78.59\pm0.08}$ & $\mathbf{90.09\pm0.06}$\\
    + MSDFA & $\mathbf{67.68\pm0.29}$ & $71.40\pm0.21$ & $\mathbf{85.89\pm0.15}$ & $71.45\pm0.26$ & $78.07\pm0.06$ & $89.88\pm0.10$ \\
         
    \bottomrule
    \end{tabular}}
    \label{tab:relative_results}
 \end{table*}

%% file: tables/sota.tex
\begin{table*}[t]
    \caption{\textit{State-of-the-art comparison.} We show that \textit{S2SD}, represented by its variants \textit{MSDF(A)}, boosts baseline objectives to state-of-the-art across literature. ($^*$) stands for Inception-V1 with frozen Batch-Norm. \textbf{Bold}: best results per literature setup. \blue{\textbf{Bluebold}}: best results per overall benchmark.}
    \setlength\tabcolsep{1.5pt}
    \footnotesize
    \centering

\resizebox{\textwidth}{!}{

 \begin{tabular}{l || c | c | c || c | c | c || c | c | c}
     \toprule
     \multicolumn{1}{l}{\textsc{Benchmarks} $\rightarrow$} & \multicolumn{3}{c}{\textsc{CUB200} \citep{cub200-2011}} & \multicolumn{3}{c}{\textsc{CARS196} \citep{cars196}} & \multicolumn{3}{c}{\textsc{SOP} \citep{lifted}}\\
     \midrule
     \textsc{Methods} $\downarrow$ & R@1 & R@2 & NMI & R@1 & R@2 & NMI & R@1 & R@10 & NMI\\
     \midrule
     \hline
     \multicolumn{10}{>{\columncolor[gray]{.8}}l}{\textbf{ResNet50-128}} \\
     \hline
     \rowcolor{vvlightgray}
     Div\&Conq \citep{Sanakoyeu_2019_CVPR}  & 65.9 & 76.6 & 69.6 & 84.6 & 90.7 & 70.3 & 75.9 & 88.4 & 90.2\\
     \rowcolor{vvlightgray}
     MIC \citep{mic}                        & 66.1 & 76.8 & 69.7 & 82.6 & 89.1 & 68.4 & 77.2 & 89.4 & 90.0\\
     \rowcolor{vvlightgray}
     PADS \citep{roth2020pads}              & 67.3 & 78.0 & 69.9 & 83.5 & 89.7 & 68.8 & 76.5 & 89.0 & 89.9\\
     \hline
     Multisimilarity+\textit{S2SD}         & 68.0  $\pm$  0.2 & 78.7 $\pm$ 0.1 & 71.7 $\pm$ 0.4 & 86.3 $\pm$ 0.1 & 91.8 $\pm$ 0.3 & 72.0 $\pm$ 0.3 & 79.0 $\pm$ 0.2 & 90.2 $\pm$ 0.1 & 90.6 $\pm$ 0.1\\
     Margin+\textit{S2SD}                  & 67.6  $\pm$  0.3 & 78.2 $\pm$ 0.2 & 70.8 $\pm$ 0.3 & 86.0 $\pm$ 0.2 & 91.8 $\pm$ 0.2 & 72.2 $\pm$ 0.2 & \textbf{80.2} $\pm$ \textbf{0.2} & \textbf{91.5} $\pm$ \textbf{0.1} & \textbf{90.9} $\pm$ \textbf{0.1}\\
     R-Margin+\textit{S2SD}                & \textbf{68.9}  $\pm$  \textbf{0.3} & \textbf{79.0} $\pm$ \textbf{0.3} & \textbf{72.1} $\pm$ \textbf{0.4} & \textbf{87.6} $\pm$ \textbf{0.2} & \textbf{92.7} $\pm$ \textbf{0.2} & \textbf{72.3} $\pm$ \textbf{0.2}& 79.2 $\pm$ 0.2 & 90.3 $\pm$ 0.1 & 90.8 $\pm$ 0.1\\
     \midrule
     \hline
     \multicolumn{10}{>{\columncolor[gray]{.8}}l}{\textbf{ResNet50-512}} \\
     \hline
     \rowcolor{vvlightgray}
     EPSHN \citep{epshn}                     & 64.9 & 75.3 &  -   & 82.7 & 89.3 &  -  & 78.3 & 90.7 &  -    \\
     \rowcolor{vvlightgray}
     NormSoft \citep{zhai2018classification} & 61.3 & 73.9 &  -   & 84.2 & 90.4 &  -   & 78.2 & 90.6 &  -    \\
     \rowcolor{vvlightgray}
     DiVA \citep{milbich2020diva}            & 69.2 & 79.3 & 71.4 & 87.6 & 92.9 & 72.2 & 79.6 & 91.2 & 90.6 \\
     \hline
     Multisimilarity+\textit{S2SD}          & 69.2 $\pm$ 0.1 & 79.1 $\pm$ 0.2 & 71.4 $\pm$ 0.2 & 89.2 $\pm$ 0.2 & 93.8 $\pm$ 0.2 & \blue{\textbf{74.0} $\pm$ \textbf{0.2}} & 80.8 $\pm$ 0.2 & \blue{\textbf{92.2} $\pm$ \textbf{0.2}} & 90.5 $\pm$ 0.3\\
     Margin+\textit{S2SD}                   & 68.8 $\pm$ 0.2 & 78.5 $\pm$ 0.2 & \blue{\textbf{72.3} $\pm$ \textbf{0.1}} & 89.3 $\pm$ 0.2 & 93.8 $\pm$ 0.2 & 73.7 $\pm$ 0.3 & \blue{\textbf{81.0} $\pm$ \textbf{0.2}} & 92.1 $\pm$ 0.2 & \textbf{91.1} $\pm$ \textbf{0.3}\\
     R-Margin+\textit{S2SD}                 & \blue{\textbf{70.1} $\pm$ \textbf{0.2}} & \blue{\textbf{79.7} $\pm$ \textbf{0.2}} & 71.6 $\pm$ 0.2 & \blue{\textbf{89.5} $\pm$ \textbf{0.2}} & \blue{\textbf{93.9} $\pm$ \textbf{0.3}} & 72.9 $\pm$ 0.3 & 80.0 $\pm$ 0.2 & 91.4 $\pm$ 0.2 & 90.8 $\pm$ 0.1\\
     \midrule
     \hline
     \multicolumn{10}{>{\columncolor[gray]{.8}}l}{\textbf{Inception-BN-512}} \\
     \hline
     \rowcolor{vvlightgray}
     DiVA \citep{milbich2020diva}     & 66.8 & 77.7 & 70.0 & 84.1 & \textbf{90.7} & 68.7 & 78.1 & \textbf{90.6} & 90.4\\
     \hline
     Multisimilarity+\textit{S2SD}   & 66.7 $\pm$ 0.3 & 77.5 $\pm$ 0.3 & \textbf{70.5} $\pm$ \textbf{0.2} & 83.8 $\pm$ 0.3 & 90.3 $\pm$ 0.2 & \textbf{69.8} $\pm$ \textbf{0.3} & \textbf{78.5} $\pm$ \textbf{0.2} & \textbf{90.6} $\pm$ \textbf{0.2} & \textbf{90.6} $\pm$ \textbf{0.1}\\
     Margin+\textit{S2SD}            & 66.8 $\pm$ 0.2 & 77.9 $\pm$ 0.2 & 69.9 $\pm$ 0.3 & \textbf{84.3} $\pm$ \textbf{0.2} & \textbf{90.7} $\pm$ \textbf{0.2} & \textbf{69.8} $\pm$ \textbf{0.2} & 78.4 $\pm$ 0.2 & 90.5 $\pm$ 0.2 & 90.4 $\pm$ 0.1\\
     R-Margin+\textit{S2SD}          & \textbf{67.4} $\pm$ \textbf{0.3} & \textbf{78.0} $\pm$ \textbf{0.4} & 70.3 $\pm$ 0.2 & 83.9 $\pm$ 0.3 & 90.3 $\pm$ 0.2 & 69.4 $\pm$ 0.2 & 78.1 $\pm$ 0.2 & 90.4 $\pm$ 0.3 & 90.3 $\pm$ 0.2\\
     \hline
     \rowcolor{vvlightgray}
     Softtriple$^*$ \citep{softriple} & 65.4 & 76.4 & 69.3 & 84.5 & 90.7 & 70.1 & 78.3 & 90.3 & \blue{\textbf{92.0}}\\
     \rowcolor{vvlightgray}
     Multisimilarity$^*$ \citep{multisimilarity} & 65.7 & 77.0 & - & 84.1 & 90.4 & -  & 78.2 & 90.5 & -   \\
     \hline
     Multisimilarity$^*$+\textit{S2SD}             & 68.2 $\pm$ 0.3 & 79.1 $\pm$ 0.2 & \textbf{71.6} $\pm$ \textbf{0.2} & 86.3 $\pm$ 0.2 & 92.2 $\pm$ 0.2 & 72.0 $\pm$ 0.3 & 78.9 $\pm$ 0.2 & 90.8 $\pm$ 0.2 & 90.6 $\pm$ 0.1\\
     Margin$^*$+\textit{S2SD}                      & 68.3 $\pm$ 0.2 & 78.8 $\pm$ 0.2 & 71.2 $\pm$ 0.2 & \textbf{87.1} $\pm$ \textbf{0.2} & \textbf{92.4} $\pm$ \textbf{0.1} & \textbf{72.2} $\pm$ \textbf{0.2} & \textbf{79.1} $\pm$ \textbf{0.2} & \textbf{91.0} $\pm$ \textbf{0.3} & 90.4 $\pm$ 0.1\\
     R-Margin$^*$+\textit{S2SD}                    & \textbf{69.6} $\pm$ \textbf{0.3} & \textbf{79.6} $\pm$ \textbf{0.3} & 71.2 $\pm$ 0.1 & 86.6 $\pm$ 0.3 & 92.1 $\pm$ 0.3 & 70.9 $\pm$ 0.2 & 78.5 $\pm$ 0.1 & 90.5 $\pm$ 0.2 & 90.0 $\pm$ 0.2\\
     \bottomrule
\end{tabular}}

\label{tab:sota}
\end{table*}

%% file: figures/metrics.tex
\begin{figure}[t]
    \centering
    \includegraphics[width=0.45\textwidth]{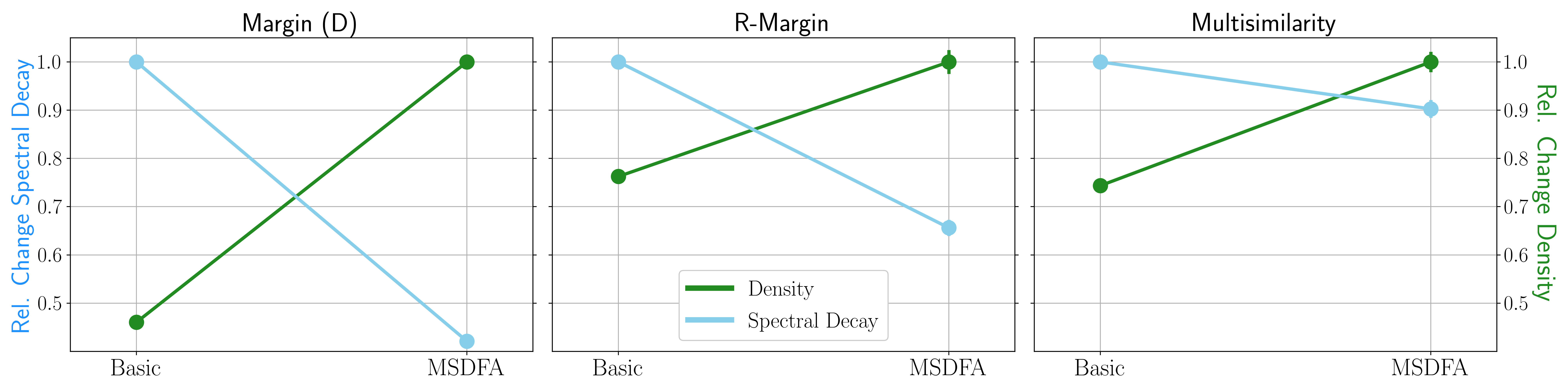}
    \caption{\textit{Generalization metrics.} \textit{S2SD} increases embedding space density and lowers spectral decay.}
    \label{fig:metrics}
\end{figure}
\vspace{-10pt}    

%% file: figures/ablations_no1.tex
\begin{figure*}[t]
    \centering
    \includegraphics[width=1\textwidth]{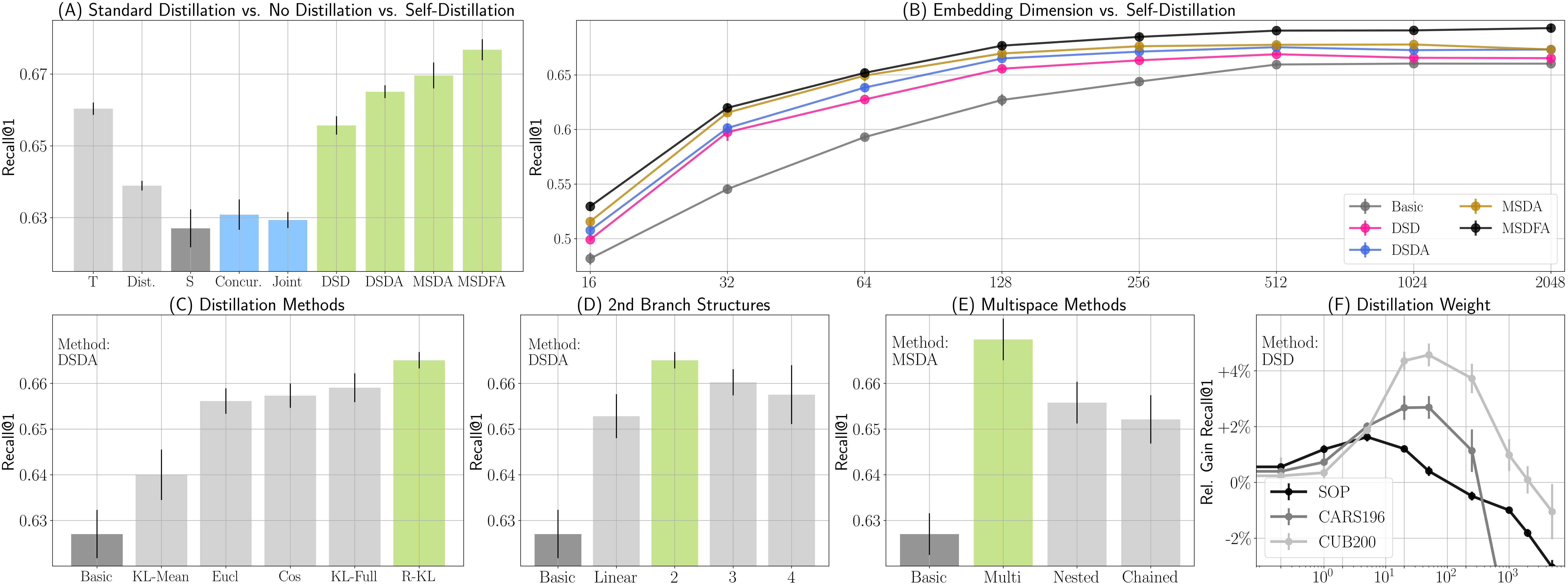}
    \caption{\textit{S2SD study and ablations.} \textbf{(A)} \textit{DSD} outperforms comparable two-stage distillation on student \textit{S} (\textit{Dist.}) using teacher (\textit{T}), with \textit{MSD(FA)} even outperforming \textit{T}. We further see that distillation is essential - training multiple spaces in parallel (\textit{Joint.}) or a detached lower-dimensional base embedding (\textit{Concur.}) gives little benefit. \textbf{(B)} We see benefits across base dimensionalities, especially in the low-dimensional regime. \textbf{(C)} We find KL-distillation between similarity vectors (\textit{R-KL}) to work best. \textbf{(D)} An additional non-linearity in aux. branches $g$ gives a boost, but going deeper hurts generalization. \textbf{(E)} Distilling each aux. embed. space (\textit{Multi}) separately compares favourable against other distillation setups s.a. \textit{Nested} and \textit{Chained} distillation. \textbf{(F)} Performance is robust to changes in weight values.}
    \label{fig:ablations_1}
\end{figure*}

%% file: appendix.tex
\onecolumn
\icmltitle{Supplementary: Simultaneous Similarity-based Self-Distillation for Deep Metric Learning}

\section{More Benchmark \& Implementation Details}\label{supp:bench_dets}
In this part, we report all relevant benchmark details omitted in the main document as well as further implementation details.\\
\subsection{Benchmarks}
\textbf{CUB200-2011} \citep{cub200-2011} contains 200 bird classes over 11,788 images, where the first and last 100 classes with 5864/5924 images are used for training and testing, respectively.\\ 
\textbf{CARS196} \citep{cars196} contains 196 car classes and 16,185 images, where again the first and last 98 classes with 8054/8131 images are used to create the training/testing split.\\ 
\textbf{Stanford Online Products (SOP)} \citep{lifted} is build around 22,634 product classes over 120,053 images and contains a provided split: 11318 selected classes with 59551 images are used for training, and 11316 classes with 60502 images for testing.
\subsection{Implementation}\label{supp:impl_dets}
We now provide further details regarding the training and testing setup utilized. For any study except the comparison against the state-of-the-art (Table \ref{tab:sota}) which uses different backbones and embedding dimensions, we follow the setup used by \cite{roth2020revisiting}\footnote{Repository: \url{github.com/Confusezius/Revisiting_Deep_Metric_Learning_PyTorch}}: This includes a ResNet50 \cite{resnet} with frozen Batch-Normalization \cite{batchnorm}, normalization of the output embeddings with dimensionality $128$ and optimization with Adam \cite{adam} using a learning rate of $10^{-5}$ and weight decay of $3\cdot10{-4}$. The input images are randomly resized and cropped from the original image size to $224\times224$ for training. Further augmentation by random horizontal flipping with $p=0.5$ is applied. During testing, center crops of size $224\times224$ are used. The batchsize is set to $112$.

Training runs on CUB200-2011 and CARS196 are done over 150 epochs and 100 epochs for SOP for all experiments without any learning rate scheduling, except for the state-of-the-art experiments (see again \ref{tab:sota}). For the latter, we made use of slightly longer training to account for conservative learning rate scheduling, which is similarly done across reference methods noted in tab. \ref{tab:sota}. Schedule and decay values are determined over validation subset performances.
All baseline DML objectives we apply our self-distillation module \textit{S2SD} on use the default parameters noted in \cite{roth2020revisiting} with the single exception of Margin Loss on SOP, where we found class margins $\beta=0.9$ to be more beneficial for distillation than the default $\beta=1.2$. This was done as changing from $\beta=1.2$ to $\beta=0.9$ had no notable impact on the baseline performance.
Finally, similar to \cite{kim2020proxy}, we found a warmup epoch of all MLPs to improve convergence on SOP. Spectral decay computations in \S\ref{sec:further_exp} follow the setting described in Supp. \ref{supp:gen_metrics}.

We implement everything in PyTorch \citep{pytorch}. Experiments are done on GPU servers containing Nvidia Titan X, P100 and T4s, however memory usage never exceeds 12GB.
Each result is averaged over five seeds, and for the sake of reproducibilty and result validity, we report mean and standard deviation, even though this is commonly neglected in DML literature.

\newpage
\section{Baseline Methods}\label{supp:base_methods}
This section provides a more detailed explanation of the DML baseline objectives we used alongside our self-distillation module \textit{S2SD} in the experimental section \ref{sec:experiments}. For additional details, we refer to the supplementary material in \cite{roth2020revisiting}. For the mathematical notation, we refer to Section \ref{sec:prelim}. We use $\psi=f\circ\phi$ to denote the feature network $\phi$ with embedding $f$, and $\psi_i$ the embedding of a sample $x_i$. Finally, alongside the method descriptions we provide the used hyperparameters.

\textbf{Margin Loss} \citep{margin} builds on triplet/pair-based losses, but introduces both class-specific, learnable boundaries $\{\beta_{y_k}\}_{k=1...C}$ (with number of classes $C$) between positive and negative pairs, as well as distance-based sampling for negatives:
\begin{align}
    \mathcal{L}_\text{margin} &= \sum_{x_i,x_j\in\mathcal{P}_\mathcal{B}} [m + (-1)^{\mathbb{I}_{y_i=y_j}}(\beta_{y_i}-d(\psi_i,\psi_j))]_+\\
    p(x_j|x_i,y_i\neq y_j) &= \min\left(\lambda, \left[ d(\psi_i,\psi_j)^{n-2}(1-\frac{1}{4}d(\psi_i,\psi_j)^2)^{\frac{n-3}{2}}\right]^{-1}\right)
\end{align}

where $\mathcal{P}_\mathcal{B}$ denotes the available pairs in minibatch $\mathcal{B}$, and $n$ the embedding dimension. Throughout this work, we use $\beta=1.2$ except for \textit{S2SD} on SOP, where we found $\beta=0.9$ to work better without changing the baseline performance. We set the learning rate for the class boundaries as $5\cdot10^{-4}$, and margin $m=0.2$.

\textbf{Regularized Margin Loss} \citep{roth2020revisiting} proposes a simple regularization scheme on the margin loss that increases the number of directions of significant variance in the embedding space by randomly exchanging a negative sample with a positive one with probability $p_\text{switch}$. For ResNet-backbones, we use $p_\text{switch}=0.4$ for CUB200, $p_\text{switch}=0.35$ for CARS196 and $p_\text{switch}=0.15$ for SOP as done in \cite{roth2020revisiting}. For Inception-based backbones, we set $p_\text{switch}=0.15$ for CUB200 and CARS196 and $p_\text{switch}=0.3$ for SOP.

\textbf{Multisimilarity Loss} \cite{multisimilarity} incorporates more similarities into training by operating directly on all positive and negative samples for an anchor $x_i$, while also incorporating a sampling operation that encourages the usage of harder training samples:
\begin{equation}
\begin{split}
    d^*_c(i,j) &= \begin{cases}
                    d_c(\psi_i,\psi_j) \qquad d_c(\psi_i,\psi_j) > \min_{j\in\mathcal{P}_i}d_c(\psi_i,\psi_j) - \epsilon\\
                    d_c(\psi_i,\psi_j) \qquad d_c(\psi_i,\psi_j) < \max_{k\in\mathcal{N}_i}d_c(\psi_i,\psi_k) + \epsilon\\
                    0\qquad\qquad\qquad\text{otherwise}
                    \end{cases}\\
\end{split}
\end{equation}
\begin{equation}
\begin{split}
    \mathcal{L}_\text{multisim} &= \frac{1}{b} \sum_{i\in \mathcal{B}} \left[ \frac{1}{\alpha}\log[1+\sum_{j\in\mathcal{P}_i}\exp(-\alpha(d^*_c(\psi_i,\psi_j)-\lambda))]\right]\\ 
    &+ \sum_{i\in \mathcal{B}} \left[\frac{1}{\beta}\log[1+\sum_{k\in\mathcal{N}_i}\exp(\beta(d^*_c(\psi_i,\psi_k)-\lambda))] \right]
\end{split}
\end{equation}

where $d_c$ denotes the cosine similarity instead of the euclidean distance, and $\mathcal{P}_i/\mathcal{N}_i$ the set of positives and negatives for $x_i$ in the minibatch, respectively. We use the default values $\alpha=2$, $\beta=40$, $\lambda=0.5$ and $\epsilon=0.1$.

\section{Evaluation Metrics}\label{supp:eval_metrics}
The evaluation metrics used throughout this work are recall @ 1 (R@1), recall @ 2 (R@2) and Normalized Mutual Information (NMI), capturing two distinct embedding space properties.

\textbf{Recall@K}, see e.g. in \cite{recall}, especially Recall@1 and Recall@2, is the primary metric used to compare the performance of DML methods and approaches, as it offers strong insights into retrieval performances of the learned embedding spaces. Given the set of embedded samples $\psi_i\in\Psi$ with $\psi_i=\psi(x_i)$ and $x_i\in\mathcal{X}$, and the sorted set of $k$ nearest neighbours for any sample $\phi_a$,
\begin{equation}\label{eq:supp_recall}
    \mathcal{F}^k_a = \minsort_{d(\phi_a,\cdot)}\argmin_{\mathcal{F}\subset\mathcal{X},|\mathcal{F}|=k} \sum_{x_f\in\mathcal{F}} d(\phi_a,\phi_f)
\end{equation}
Recall@K is measured as
\begin{equation}
    \text{Recall@K} = \frac{1}{|\mathcal{X}|}\sum_{x_i\in\mathcal{X}} 
    \begin{cases}
    1\quad\exists x_k\in\mathcal{F}^k_i \text{s.t.} y_k=y_i\\
    0\quad\text{otherwise}
    \end{cases}
\end{equation}
which evaluates how likely semantically corresponding pairs (as determined here by the labelling $y_i\in\mathcal{Y}$) will occur in a neighbourhood of size $k$.

\textbf{Normalized Mutual Information (NMI)}, see \cite{nmi}, evaluates the clustering quality of the embedded samples $\Psi$ (taken from $\mathcal{X}$). It is computed by first clustering with $K$ cluster centers, usually corresponding to the number of classes available, using a cluster method of choice s.a. K-Means \citep{kmeans}. This assigns each sample $x_i$ a cluster label/id $\omega_i$ based on the nearest cluster centroid. With $\eta_k=\{i|\omega_i=\omega_k\}$ the set of samples with cluster label, $\Omega=\{\eta_k\}_k^K$ the set of cluster sets, $\nu_k=\{i|y_i=y_k\}$ the set of samples with true label $y_k$ and $\Upsilon=\{\nu_k\}_k^K$ the set of class label sets, the Normalized Mutual Information is given as
\begin{equation}
    \text{NMI}(\Omega,\Upsilon) = \frac{I(\Omega,\Upsilon)}{2\cdot\left(H(\Omega) + H(\Upsilon)\right)}
\end{equation}
with mutual information $I(\cdot,\cdot)$ and entropy $H(\cdot)$.

\section{Generalization Metrics}\label{supp:gen_metrics}
\textbf{Embedding Space Density.} Given sets of embeddings $\Psi$, we first define the average inter-class distance as
\begin{equation}
    \pi_\text{inter}(\Psi) = \frac{1}{Z_\text{inter}}\sum_{y_l,y_k,l\neq k}d(\mu(\Psi_{y_l}),\mu(\Psi_{y_k}))
\end{equation}
which measures the average distances between groups of embeddings with respective classes $y_l$ and $y_k$, estimated by the respective class centers $\mu(\cdot)$. $Z_\text{inter}$ denotes a normalization constant based on the number of available classes. We also introduce the average intra-class distance as the mean distance between samples within their respective class
\begin{equation}
    \pi_\text{intra}(\Psi) = \frac{1}{Z_\text{intra}}\sum_{y_l\in\mathcal{Y}}\sum_{\psi_i,\psi_j\in\Psi_{y_l}, i\neq j} d(\psi_i,\psi_j)
\end{equation}
again with normalization constant $Z_\text{intra}$ and set of embeddings with class $y_l$, $\Psi_{y_l}$. Given these two quantities, the embedding space density is then defined as 
\begin{equation}
    \pi_\text{ratio}(\Psi) = \frac{\pi_\text{intra}(\Psi)}{\pi_\text{inter}(\Psi)}
\end{equation}
and effectively measured how densely samples and classes are grouped together. \cite{roth2020revisiting} show that optimizing the DML problem while keeping the embedding space density high, i.e. without aggressive clustering, encourages better generalization to unseen test classes.

\textbf{Spectral Decay.}
The spectral decay metric $\rho(\Psi)$ defines the KL-divergence between the (sorted) spectrum of $D$ singular values $\mathcal{S}^\text{singular}_\Psi$ (obtained via Singular Value Decomposition (SVD)) and a $D$-dimensional uniform distribution $\mathcal{U}_D$, and is inversly related to the entropy of the embedding space:
\begin{equation}
    \rho(\Psi) = \mathcal{D}_\text{KL}\left(\mathcal{U}_D,\mathcal{S}^\text{singular}_\Psi\right)
\end{equation}
It does not account for class distributions. \cite{roth2020revisiting} show that doing DML while encouraging a high-entropy feature space notably benefits the generalization performance. In our experiments, we disregard the first 10 singular vectors (out of 128) to
highlight the feature diversity. This is important, as we evaluate the spectral decay within the same objectives, which results in the first few singular values to be highly similar.

\newpage
\section{Additional Experiments}\label{supp:add_exps}
This part extends the set of ablations experiments performed in section \ref{sec:ablations} in the main paper.\\
\textbf{a. Detaching target spaces for distillation.} We examine whether it is preferable to detach the target embeddings from the distillation loss (see eq. \ref{eq:kl_distill}), as we want the reference embedding space to approximate the higher-dimensional relations. Similarly, we do not want the target embedding networks $g_i$ to reduce high-dimensional to lower-dimensional relations to optimizer for the distillation constraint. As can be seen in fig \ref{fig:ablations_2}C, it is indeed the case that detaching the target embedding spaces is notably beneficial for a stronger reference embedding, supporting the previous motivation.\\
\textbf{b. Influence of varying target dimensions.} As noted at the beginning of section \ref{sec:experiments}, we set the target dimension for dual self-distillation (\textbf{DSD}) to $d=2048$, which we motivate through a small ablation study in fig. \ref{fig:ablations_2}A, with \textit{TD} denoting the target dimension of choice. As can be seen, benefits plateau when the target dimension reaches more than four times the reference dimension. However, to be directly comparable to high-dimensional reference settings, we set $d=2048$ as default.\\
\textbf{c. Ablating multiple distillation scales.} Going further, we extend the module with additional embedding branches to the multiscale self-distillation approach (\textbf{MSD}), all operating in different, but higher-than-reference dimension. As already shown in Figure \ref{fig:ablations_1}B in the main paper, there is a benefit of multiscale distillations by encouraging reusable sample relations. In this part, we motivate the choice of four target branches (as noted in sec. \ref{sec:experiments}). Looking at figure \ref{fig:ablations_2}A, where $B$ denotes the number of additional target spaces, we can see a benefit in multiple additional target spaces of ascending dimension. As the improvements saturate after $B=4$, we simply set this as the default value. However, the additional benefits of going to multiscale from dual distillation are not as high as going from no to dual target space distillation, showcasing the general benefit of high-dimensional concurrent self-distillation. Finally, we highlight that a multiscale approach slightly outperforms a multibranch distillation setup (Fig. \ref{fig:ablations_2}A, \textit{Multi-B}) where each target branch has the same target dimension of $2048$ while introducing less additional parameters.\\
\textbf{d. Finer-grained feature distillation.} As already shown in section \ref{sec:experiments} and again in figure \ref{fig:ablations_2}B, we see benefits of feature distillation, using the (globally averaged) normalized penultimate feature space. It therefore makes sense to investigate the benefits of distilling even more fine-grained feature representation. Defining $\mathcal{P}=[(3,3),(1,1),(2,2),(4,1)]$ as the pooling window size applied to the non-average penultimate feature representation, we investigate less compressed feature representation space. As can be seen in fig. \ref{fig:ablations_2}B, where $P$ denotes the index to $\mathcal{P}$, there appears to be no benefits in distilling feature representations higher up the network.\\ 
\textbf{e. Runtime comparison of base dimensionalities.} We highlight relative retrieval times at different base dimensionalities in Tab. \ref{tab:test_time} using \textsf{faiss} \citep{faiss} on a NVIDIA 1080Ti and a synthetic set of $N=250000$ embeddings of dimensionality $d\in[32,64,128,256,512,1024,2048]$. With \textit{S2SD} matching $d=64/128$ to base dimensionalities $d=512/2048$ (see \S\ref{sec:further_exp}), runtime can be reduced by up to a magnitude.

\input{figures/ablations_no2}

\begin{table}[h!]
    \centering
    \resizebox{\textwidth}{!}{
    \begin{tabular}{l|c|c|c|c|c|c|c}
    \toprule
         \textbf{Dimensionality} $d$ & 32   &  64  &  128 &  256 &  512 &  1024 & 2048  \\
         \hline
         \textbf{Runtime} (\textit{s})        & 1.54$\pm$0.00 & 1.98$\pm$0.00 & 2.71$\pm$0.00 & 4.35$\pm$0.00 & 7.38$\pm$0.01 & 13.83$\pm$0.02 & 27.21$\pm$0.17 \\
    \bottomrule
    \end{tabular}}
    \caption{\small{Sample retrieval times for 250000 embeddings with varying base dimensionalities.}}
    \label{tab:test_time}
\end{table}

\newpage
\section{Pseudo-Code}\label{supp:pseudo_code}
\begin{lstlisting}[language=Python, morekeywords={self,nn,norm,F}, caption=PyTorch Implementation for \textit{S2SD}.]
import torch, torch.nn as nn, torch.nn.functional as F
from F import normalize as norm

"""
Parameters:
    self.base_criterion: base DML objective
    self.trgt_criteria: list of DML objectives for target spaces
    self.trgt_nets: Module list of auxiliary embedding MLPs
    self.dist_gamma: distillation weight
    self.it_before_feat_distill: iterations before feature distill  
"""

def forward(self, batch, labels, pre_batch, **kwargs):
    """
    Args:
        batch: image embeddings, shape: bs x d
        labels: image labels, shape: bs
        pre_batch: penultimate network features, shape: bs x d*
    """
    bs, batch = len(batch), norm(batch, dim=-1)

    ### Compute ref. sample relations and loss on ref. embedding space
    base_smat = batch.mm(batch.T)
    base_loss = self.base_criterion(batch, labels, **kwargs)

    ### Do global average pooling (and max. pool if wanted)
    avg_pre_batch  = nn.AdaptiveAvgPool2d(1)(pre_batch).view(bs,-1)
    avg_pre_batch += nn.AdaptiveMaxPool2d(1)(pre_batch).view(bs,-1)

    ### Computing MSDA loss (Targets & Distillations)
    dist_losses, trgt_losses  = [], []
    for trgt_crit,trgt_net in zip(self.trgt_criteria,self.trgt_nets):
        trgt_batch     = norm(trgt_net(avg_pre_batch),dim=-1)
        trgt_loss      = trgt_crit(trgt_batch, labels, **kwargs)
        trgt_smat      = trgt_batch.mm(trgt_batch.T)
        base_trgt_dist = self.kl_div(base_smat, trgt_smat.detach())
        trgt_losses.append(trgt_loss)
        dist_losses.append(base_trgt_dist)

    ### MSDA loss
    multi_dist_loss  = (base_loss+torch.stack(trgt_losses).mean())/2.
    multi_dist_loss += self.dist_gamma*torch.stack(dist_losses).mean()

    ### Distillation of penultimate features -> MSDFA 
    src_feat_dist = 0
    if self.it_count>=self.it_before_feat_distill:
        n_avg_pre_batch = norm(avg_pre_batch, dim=-1).detach()
        avg_feat_smat   = n_avg_pre_batch.mm(n_avg_pre_batch.T)
        src_feat_dist   = self.kl_div(base_smat, avg_feat_smat.detach())

    ### Total S2SD training objective 
    total_loss = multi_distill_loss + self.dist_gamma*src_feat_dist
    self.it_count+=1
    return total_loss

def kl_div(self, A, B, T=1):
    log_p_A = F.log_softmax(A/self.T, dim=-1)
    p_B     = F.softmax(B/self.T, dim=-1)
    kl_d    = F.kl_div(log_p_A, p_B,reduction='sum')*T**2/A.size(0)
    return kl_d
\end{lstlisting}

\newpage
\section{Detailed Evaluation Results}\label{supp:detailed_1}
This table contains all method ablations for a fair evaluation as used in Section \ref{sec:comp} and Table \ref{tab:relative_results}.
\input{tables/Full_Results}

\section{Evaluation Results using mAP@R}\label{supp:detailed_map}
This table measures performance of methods investigated in Table \ref{tab:relative_results} using the mAP@R(@1000) metric used in \cite{roth2020revisiting}. The results here coincide with those measured using Recall@1. This comes at no surprise, as both metrics are strongly correlated when measuring the performance of Deep Metric Learning methods \citep{roth2020revisiting}.
\input{tables/Full_Results_mAP}

\section{Detailed Ablation Results}\label{supp:detailed_2}
Detailed values to the ablation experiments done in section \ref{sec:ablations} and \ref{supp:add_exps}.
\input{tables/ablations}

%% file: figures/ablations_no2.tex
\begin{figure}[t]
\centering
\includegraphics[width=1\linewidth]{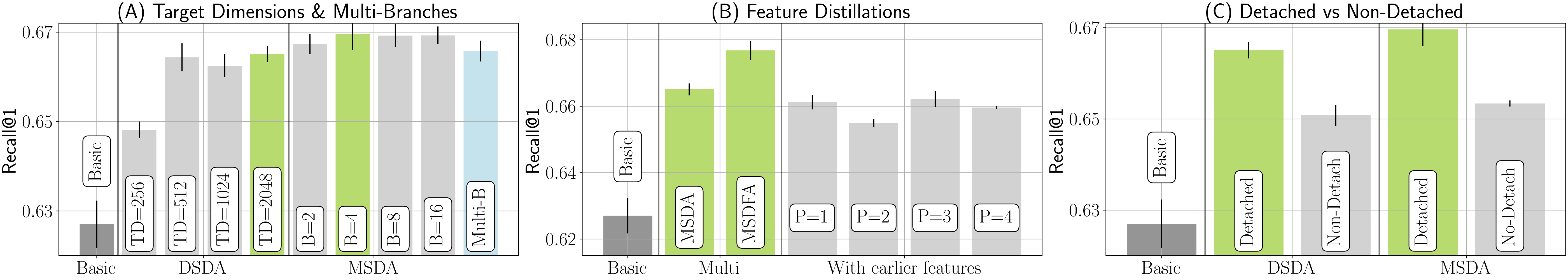}
\caption{\small{\textit{Additional ablations.} \textbf{(A)} Increasing target dimensions offers notable improvements. We opt for a target dimension of 2048 due to slightly higher mean improvements. For multiple embedding branches (\#B), there seems to be an optimum at four branches. \textbf{(B)} Furthermore, feature distillation gives another notable boost. However, this only holds for the globally averaged penultimate feature representation. When distilling more fine-grained feature representations, performance degenerates (where \#P denotes smaller pooling windows applied to the penultimate feature representation). \textbf{(C)} We show that detached auxiliary branches for distillation are crucial to higher improvements, as we want the reference embedding space to approximate the higher-dimensional one.}}
\label{fig:ablations_2}
\end{figure}


%% file: tables/Full_Results.tex
\begin{table}[h]
\caption{\textit{Detailed Comparison of Recall@1 and NMI performances against well performing DML objectives examined in section \ref{sec:comp}.} This is the complete version to table \ref{tab:relative_results}. All results are computed over 5-run averages. ($^*$) For Margin Loss and SOP, we found $\beta=0.9$ to give better distillation results without notably influencing baseline performance.}
 \footnotesize
   \setlength\tabcolsep{1.4pt}
   \centering
   \begin{tabular}{l|c|c||c|c||c|c}
     \toprule
     \multicolumn{1}{l}{\textbf{Benchmarks}$\rightarrow$} & \multicolumn{2}{c}{\textsc{CUB200-2011}} & \multicolumn{2}{c}{\textsc{CARS196}} & \multicolumn{2}{c}{\textsc{SOP}} \\
     \midrule
     \textbf{Approaches} $\downarrow$ & R@1 & NMI & R@1 & NMI & R@1 & NMI\\
    \midrule
    \textbf{Margin}($^*$) & $63.09\pm0.46$ & $68.21\pm0.33$ & $79.86\pm0.33$ & $67.36\pm0.34$ & $78.43\pm0.07$ & $90.40\pm0.03$\\        
    + DSD & $65.11\pm0.18$ & $69.65\pm0.44$ & $83.19\pm0.18$ & $69.28\pm0.56$ & $79.05\pm0.12$ & $90.52\pm0.18$\\
    + DSDA & $65.77\pm0.55$ & $69.85\pm0.25$ & $83.92\pm0.08$ & $69.95\pm0.21$ & $77.78\pm0.15$ & $90.29\pm0.08$\\
    + MSD & $66.13\pm0.34$ & $70.83\pm0.27$ & $83.63\pm0.31$ & $69.80\pm0.36$ & $79.26\pm0.15$ & $90.60\pm0.10$\\
    + MSDA & $66.14\pm0.32$ & $70.82\pm0.18$ & $84.31\pm0.12$ & $70.17\pm0.30$ & $78.04\pm0.11$ & $90.45\pm0.05$\\
    + MSDF & $67.58\pm0.32$ & $71,47\pm0.19$ & $85.55\pm0.23$ & $71.68\pm0.54$ & $79.63\pm0.14$ & $90.70\pm0.09$\\
    + MSDFA & $67.21\pm0.23$ & $71.43\pm0.25$ & $86.45\pm0.35$ & $71.46\pm0.24$ & $78.82\pm0.09$ & $90.49\pm0.06$\\
    
    \midrule
    \textbf{R-Margin} & $64.93\pm0.42$ & $68.36\pm0.32$ & $82.37\pm0.13$ & $68.66\pm0.47$ & $77.58\pm0.11$ & $90.42\pm0.03$\\ 
    + DSD & $66.58\pm0.08$ & $70.03\pm0.41$ & $84.64\pm0.16$ & $70.87\pm0.18$ & $77.86\pm0.10$ & $90.50\pm0.03$\\
    + DSDA & $67.11\pm0.43$ & $70.39\pm0.48$ & $84.32\pm0.36$ & $70.85\pm0.16$ & $77.79\pm0.11$ & $90.37\pm0.04$\\
    + MSD & $66.81\pm0.27$ & $70.47\pm0.16$ & $85.01\pm0.10$ & $71.67\pm0.40$ & $78.00\pm0.06$ & $90.47\pm0.04$\\
    + MSDA & $67.31\pm0.41$ & $71.01\pm0.24$ & $85.34\pm0.17$ & $71.85\pm0.20$ & $77.93\pm0.06$ & $90.29\pm0.08$\\
    + MSDF & $68.12\pm0.30$ & $71.80\pm0.33$ & $85.78\pm0.22$ & $72.24\pm0.31$ & $78.57\pm0.09$ & $90.58\pm0.02$\\
    + MSDFA & $68.58\pm0.26$ & $71.64\pm0.40$ & $86.81\pm0.35$ & $71.48\pm0.29$ & $78.00\pm0.11$ & $90.41\pm0.02$\\
    \midrule
    \textbf{Multisimilarity} & $62.80\pm0.70$ & $68.55\pm0.38$ & $81.68\pm0.19$ & $69.43\pm0.38$ & $77.99\pm0.09$ & $90.00\pm0.02$\\
    + DSD & $65.57\pm0.26$ & $70.08\pm0.33$ & $83.51\pm0.20$ & $70.30\pm0.05$ & $78.23\pm0.04$& $90.08\pm0.04$\\
    + DSDA & $66.60\pm0.43$ & $70.74\pm0.40$ & $84.42\pm0.28$ & $70.36\pm0.34$ & $77.92\pm0.12$ & $89.99\pm0.04$\\
    + MSD & $65.80\pm0.16$ & $70.53\pm0.01$ & $83.98\pm0.10$ & $71.34\pm0.09$ & $78.42\pm0.09$ & $90.09\pm0.03$\\
    + MSDA & $66.96\pm0.36$ & $70.77\pm0.05$ & $85.04\pm0.14$ & $71.09\pm0.23$ & $77.98\pm0.05$ & $90.02\pm0.04$\\
    + MSDF & $67.04\pm0.29$ & $71.87\pm0.19$ & $85.69\pm0.19$ & $72.77\pm0.13$ & $78.59\pm0.08$ & $90.09\pm0.06$\\
    + MSDFA & $67.68\pm0.29$ & $71.40\pm0.21$ & $85.89\pm0.15$ & $71.45\pm0.26$ & $78.07\pm0.06$ & $89.88\pm0.10$\\
         
    \bottomrule
    \end{tabular}
    \label{tab:relative_results_long}
 \end{table}

%% file: tables/Full_Results_mAP.tex
\begin{table}[h]
\caption{\textit{Detailed Comparison of mAP@R} (as used in \cite{roth2020revisiting} and \cite{musgrave2020metric} and based on the formulation used in \cite{roth2020revisiting}) against well performing DML objectives examined in section \ref{sec:comp}.. All results are computed over 5-run averages. ($^*$) For Margin Loss and SOP, we found $\beta=0.9$ to give better distillation results without notably influencing baseline performance. \textbf{Bold} denotes best results per objective and dataset. \blue{\textbf{Bluebold}} denotes best performance per dataset.}
 \footnotesize
   \setlength\tabcolsep{1.4pt}
   \centering
   \begin{tabular}{l|c||c||c}
     \toprule
     \multicolumn{1}{l}{\textbf{Benchmarks}$\rightarrow$} & \multicolumn{1}{c}{\textsc{CUB200-2011}} & \multicolumn{1}{c}{\textsc{CARS196}} & \multicolumn{1}{c}{\textsc{SOP}} \\
     \midrule
     \textbf{Approaches} $\downarrow$ & mAP & mAP & mAP\\
    \midrule
    \rowcolor{vvlightgray}
    \textbf{Margin}($^*$) & $32.63\pm0.40$ & $32.50\pm0.28$ & $46.90\pm0.16$ \\        
    + DSD & $33.85\pm0.38$ & $34.01\pm0.39$ & $47.39\pm0.18$\\
    + MSD & $34.79\pm0.35$ & $34.64\pm0.31$ & $48.17\pm0.07$\\
    + MSDF & $35.68\pm0.29$ & $35.26\pm0.41$ & \blue{$\mathbf{48.24}\pm0.10$}\\
    + MSDFA & $\mathbf{35.98}\pm0.23$ & $\mathbf{35.98}\pm0.40$ & $47.04\pm0.26$\\
    \midrule
    \rowcolor{vvlightgray}
    \textbf{R-Margin} & $33.38\pm0.27$ & $34.57\pm0.30$ & $46.02\pm0.14$\\ 
    + DSD & $34.46\pm0.30$ & $35.12\pm0.22$ & $46.20\pm0.19$\\
    + MSD & $35.11\pm0.41$ & $35.78\pm0.40$ & $46.59\pm0.16$\\
    + MSDF & $35.99\pm0.36$ & $37.32\pm0.40$ & $\mathbf{47.08}\pm0.17$\\
    + MSDFA & \blue{$\mathbf{36.25}\pm0.37$} & \blue{$\mathbf{37.67}\pm0.35$} & $46.71\pm0.16$\\
    \midrule
    \rowcolor{vvlightgray}
    \textbf{Multisimilarity} & $30.92\pm0.49$ & $31.92\pm0.44$ & $46.23\pm0.08$\\
    + DSD & $33.20\pm0.34$ & $33.67\pm0.27$ & $46.21\pm0.15$\\
    + MSD & $34.00\pm0.35$ & $34.67\pm0.26$ & $46.45\pm0.11$\\
    + MSDF & $35.16\pm0.32$ & $\mathbf{35.52}\pm0.51$ & $\mathbf{46.52}\pm0.17$\\
    + MSDFA & $\mathbf{35.35}\pm0.24$ & $35.13\pm0.35$ & $45.39\pm0.28$\\

    \bottomrule
    \end{tabular}
    \label{tab:relative_results_map}
 \end{table}

\begin{table}[ht!]
\centering
\caption{Additional ProxyAnchor \citep{kim2020proxy} results with and without \textit{S2SD} variants using the proposed, but different, default architecture in \citep{kim2020proxy} to highlight that \textit{S2SD} works equally well on already strong proxy-based objectives objectives with different architectural settings as well.}
\centering
\resizebox{0.95\textwidth}{!}{   
\begin{tabular}{l||c|c||c|c||c|c}
 \toprule
 \multicolumn{1}{l}{\textbf{Benchmarks}$\rightarrow$} & \multicolumn{2}{c}{\textsc{CUB200-2011}} & \multicolumn{2}{c}{\textsc{CARS196}} & \multicolumn{2}{c}{\textsc{SOP}} \\
\midrule
 \textbf{Setting} & R@1 & NMI & R@1 & NMI & R@1 & NMI \\
\midrule     
\rowcolor{vvlightgray}
\textbf{ProxyAnchor} & $64.58\pm0.23$ & $68.95\pm0.24$ & $82.55\pm0.41$ & $69.49\pm0.30$ & $78.33\pm0.08$ & $90.24\pm0.06$\\    
+ DSD & $65.50\pm0.47$ & $69.97\pm0.55$ & $83.52\pm0.11$ & $70.76\pm0.17$ & $78.33\pm0.08$ & $90.24\pm0.06$\\
+ MSD & $65.92\pm0.28$ & $69.88\pm0.21$ & $83.99\pm0.33$ & $70.95\pm0.19$ & $78.47\pm0.03$ & $90.29\pm0.06$\\
+ MSDF & $66.71\pm0.12$ & $70.60\pm0.24$ & $85.20\pm0.09$ & $71.19\pm0.18$ & $78.50\pm0.04$ & $90.31\pm0.03$\\
\end{tabular}}
\label{tab:proxyanchor_results}
\end{table}

%% file: tables/ablations.tex



\begin{table}[h]
 \small
   \centering
    \caption{\textit{Experiment: Comparison of concurrent self-distillation against standard 2-stage distillation.} This table also shows that training without distillation (\textit{Joint}) or training in high dimension while learning a detached low-dimensional embedding layer (\textit{Concur.}) does not benefit performance notably. See fig. \ref{fig:ablations_1}A. All results are computed over 5-run averages.}   
   \begin{tabular}{l||c|c}
     \toprule
     Experiment & Setting & R@1\\
     \midrule
     \multirow{9}{*}{Distillation} & Best Teacher (d=1024) & $66.04\pm0.17$\\
     & Base Student (d=128) & $62.70\pm0.53$\\
     & Distill Student (d=128) & $63.89\pm0.14$\\
     & Concur. Student (d=128) & $63.08\pm0.42$\\
     & Joint Student (d=128) & $62.93\pm0.22$\\
     & DSD   (d=128) & $65.57\pm0.26$\\
     & DSDA  (d=128) & $66.51\pm0.18$\\
     & MSDA (d=128) & $66.96\pm0.36$\\
     & MSDFA (d=128) & $67.68\pm0.29$\\
    \bottomrule
    \end{tabular}
    \label{tab:ablations_6}
 \end{table}

 \begin{table}[h]
 \small
   \centering
    \caption{\textit{Experiment: Benefit of self-distillation across embedding dimensionalities.} These results go along with \ref{fig:ablations_1}B. All results are computed over 5-run averages.}   
   \begin{tabular}{l||c|c|c|c}
     \toprule
     Experiment & Setting & R@1 & Setting & R@1\\
     \midrule
\multirow{28}{*}{Embedding Dimensionality} & Base (d=16) & $48.18\pm0.54$& MSD (d=256) & $66.90\pm0.11$\\
 & Basic (d=32) & $54.54\pm0.42$& MSD (d=512) & $67.07\pm0.02$\\     
 & Basic (d=64) & $59.31\pm0.26$& MSD (d=1024) & $66.69\pm0.11$\\ 
 & Basic (d=128) & $62.70\pm0.53$& MSD (d=2048) & $66.68\pm0.18$\\ 
 & Basic (d=256) & $64.39\pm0.30$& MSDA (d=16) & $51.57\pm0.39$\\ 
 & Basic (d=512) & $65.95\pm0.19$& MSDA (d=32) & $61.55\pm0.23$\\ 
 & Basic (d=1024) & $66.04\pm0.17$& MSDA (d=64) & $64.94\pm0.50$\\ 
 & Basic (d=2048) & $66.03\pm0.20$& MSDA (d=128) & $66.96\pm0.36$\\ 
 & DSD (d=16) & $49.92\pm0.09$& MSDA (d=256) & $67.63\pm0.34$\\ 
 & DSD (d=32) & $59.75\pm0.78$& MSDA (d=512) & $67.76\pm0.26$\\ 
 & DSD (d=64) & $62.75\pm0.15$& MSDA (d=1024) & $67.79\pm0.10$\\ 
 & DSD (d=128) & $65.57\pm0.26$& MSDA (d=2048) & $67.33\pm0.09$\\ 
 & DSD (d=256) & $66.34\pm0.07$& MSDF (d=16) & $51.99\pm0.57$\\ 
 & DSD (d=512) & $66.89\pm0.15$& MSDF (d=32) & $61.61\pm0.41$\\ 
 & DSD (d=1024) & $66.57\pm0.11$& MSDF (d=64) & $65.31\pm0.23$\\ 
 & DSD (d=2048) & $66.54\pm0.28$& MSDF (d=128) & $66.66\pm0.29$\\ 
 & DSDA (d=16) & $50.77\pm0.71$& MSDF (d=256) & $67.47\pm0.11$\\ 
 & DSDA (d=32) & $60.13\pm0.45$& MSDF (d=512) & $67.59\pm0.03$\\ 
 & DSDA (d=64) & $63.84\pm0.36$& MSDF (d=1024) & $67.40\pm0.15$\\ 
 & DSDA (d=128) & $66.51\pm0.18$& MSDF (d=2048) & $67.01\pm0.35$\\ 
 & DSDA (d=256) & $67.13\pm0.24$& MSDFA (d=16) & $52.96\pm0.44$\\ 
 & DSDA (d=512) & $67.54\pm0.24$& MSDFA (d=32) & $61.98\pm0.36$\\ 
 & DSDA (d=1024) & $67.27\pm0.22$& MSDFA (d=64) & $65.19\pm0.40$\\ 
 & DSDA (d=2048) & $67.33\pm0.30$& MSDFA (d=128) & $67.68\pm0.29$\\ 
 & MSD (d=16) & $50.51\pm0.21$& MSDFA (d=256) & $68.48\pm0.28$\\ 
 & MSD (d=32) & $60.00\pm0.28$& MSDFA (d=512) & $69.06\pm0.14$\\ 
 & MSD (d=64) & $63.74\pm0.24$& MSDFA (d=1024) & $69.08\pm0.22$\\ 
 & MSD (d=128) & $65.80\pm0.16$& MSDFA (d=2048) & $69.29\pm0.35$\\ 
    \bottomrule
    \end{tabular}
    \label{tab:ablations_5}
 \end{table}

 \begin{table}[h]
 \small
   \centering
    \caption{\textit{Experiment: Methods of distillation between reference and target embedding spaces.} See fig. \ref{fig:ablations_1}C. Used Method: \textbf{DSDA}. All results are computed over 5-run averages.}   
   \begin{tabular}{l||c|c}
     \toprule
     Experiment & Setting & R@1\\
     \midrule
\multirow{7}{*}{Distillation Methods} & R-KL & $66.51\pm0.18$\\
 & Cos & $65.73\pm0.27$\\
 & Eucl & $65.61\pm0.28$\\
 & KL-Full & $65.91\pm0.32$\\
 & KL-Mean & $64.00\pm0.55$\\
 & Basic & $62.70\pm0.53$\\
    \bottomrule
    \end{tabular}
    \label{tab:ablations_3}
 \end{table}

\begin{table}[h]
 \small
   \centering
    \caption{\textit{Experiment: Structure of the secondary branch.} More specifically, this table contains specific values used in fig. \ref{fig:ablations_1}D. Used Method: \textbf{DSDA}. All results are computed over 5-run averages.}   
   \begin{tabular}{l||c|c}
     \toprule
     Experiment & Setting & R@1\\
     \midrule
\multirow{5}{*}{Secondary Branch Structure} & 2 Layers & $66.51\pm0.18$\\
 & 3 Layers & $66.03\pm0.29$\\
 & 4 Layers & $65.76\pm0.65$\\
 & Linear & $65.28\pm0.48$\\
 & Basic & $62.70\pm0.53$\\
    \bottomrule
    \end{tabular}
    \label{tab:ablations_2}
 \end{table}

 \begin{table}[h]
 \small
   \centering
    \caption{\textit{Experiment: Different distillation hierarchies.} See fig. \ref{fig:ablations_1}E. Used Method: \textbf{MSDA}. All results are computed over 5-run averages.}   
   \begin{tabular}{l||c|c}
     \toprule
     Experiment & Setting & R@1\\
     \midrule
\multirow{4}{*}{Distillation Hierarchies} & Basic & $62.70\pm0.53$\\
 & Straight & $66.96\pm0.36$\\
 & Fully & $65.58\pm0.46$\\
 & Stacked & $65.21\pm0.25$\\
    \bottomrule
    \end{tabular}
    \label{tab:ablations_10}
 \end{table}

 \begin{table}[h]
 \small
   \centering
    \caption{\textit{Experiment: Influence of distillation weight $\gamma$.} See fig. \ref{fig:ablations_1}F. Used Method: \textbf{DSD}. All results are computed over 5-run averages.}   
   \begin{tabular}{l||c|c|c|c}
     \toprule
     Experiment & Setting & R@1 CUB200 & R@1 CARS196 & R@1 SOP\\
     \midrule
    \multirow{10}{*}{Weight Ablation} & 0.0 & $62.70\pm0.53$ & $81.32\pm0.36$ & $77.78\pm0.06$\\
     & 0.2 & $62.85\pm0.41$ & $81.65\pm0.40$ & $78.22\pm0.07$\\
     & 1.0 & $62.92\pm0.16$ & $81.92\pm0.51$ & $78.71\pm0.10$\\
     & 5.0 & $63.88\pm0.19$ & $82.96\pm0.10$ & $79.05\pm0.12$\\
     & 20.0 & $65.43\pm0.21$ & $83.50\pm0.35$ & $78.72\pm0.10$\\
     & 50.0 & $65.57\pm0.26$ & $83.51\pm0.33$ & $78.10\pm0.13$\\
     & 250.0 & $65.04\pm0.33$ & $82.25\pm0.62$ & $77.41\pm0.12$\\
     & 1000.0 & $63.32\pm0.36$ & $77.00\pm0.66$ & $77.01\pm0.08$\\
     & 2000.0 & $62.76\pm0.31$ & $72.72\pm0.85$ & $76.37\pm0.11$\\
     & 5000.0 & $62.05\pm0.62$ & $70.90\pm0.97$ & $75.42\pm0.20$\\  

    \bottomrule
    \end{tabular}
    \label{tab:ablations_4}
 \end{table}

 \begin{table}[h]
 \small
   \centering
    \caption{\textit{Experiment: Evaluation target dimensions and levels of multiscale distillation.} See fig. \ref{fig:ablations_2}A. All results are computed over 5-run averages.}   
   \begin{tabular}{l||c|c}
     \toprule
     Experiment & Setting & R@1\\
     \midrule
\multirow{5}{*}{Target Dimensionalities} & Basic & $62.70\pm0.53$\\
 & DSDA, TD=256 & $64.82\pm0.18$\\
 & DSDA, TD=512 & $66.44\pm0.31$\\
 & DSDA, TD=1024 & $66.25\pm0.26$\\
 & DSDA, TD=2048 & $66.51\pm0.18$\\
\multirow{4}{*}{MultiScale distillation} & MSDA, \#B=2 & $66.76\pm0.23$\\
 & MSDA, \#B=4 & $66.96\pm0.36$\\
 & MSDA, \#B=8 & $66.72\pm0.25$\\
 & MSDA, \#B=16 & $66.59\pm0.20$\\
    \bottomrule
    \end{tabular}
    \label{tab:ablations_7}
 \end{table}

 \begin{table}[h]
 \small
   \centering
    \caption{\textit{Experiment: Is it beneficial to distill more fine-grained features?} See fig. \ref{fig:ablations_2}B. All results are computed over 5-run averages.}
    \label{tab:ablations_8}   
   \begin{tabular}{l||c|c}
     \toprule
     Experiment & Setting & R@1\\
     \midrule
\multirow{7}{*}{Earlier Features} & Basic & $62.70\pm0.53$\\
 & MSDA & $66.96\pm0.36$\\
 & MSDFA & $67.68\pm0.29$\\
 & MSDFA, \#P=1 & $66.13\pm0.22$\\
 & MSDFA, \#P=2 & $65.49\pm0.12$\\
 & MSDFA, \#P=3 & $66.22\pm0.24$\\
 & MSDFA, \#P=4 & $65.96\pm0.04$\\
    \bottomrule
    \end{tabular}
 \end{table}

 \begin{table}[h]
 \small
   \centering
    \caption{\textit{Experiment: Is it necessary to detach auxiliary branches for distillation?} See fig. \ref{fig:ablations_2}C. All results are computed over 5-run averages.}   
   \begin{tabular}{l||c|c}
     \toprule
     Experiment & Setting & R@1\\
     \midrule
\multirow{5}{*}{Branch Detaching} & Basic & $62.70\pm0.53$\\
 & DSDA, Detached & $66.51\pm0.18$\\
 & DSDA, Non-Detach & $65.08\pm0.23$\\
 & MSDA, Detached & $66.96\pm0.36$\\
 & MSDA, No-Detach & $65.34\pm0.07$\\
    \bottomrule
    \end{tabular}
    \label{tab:ablations_9}
 \end{table}